%% file: arxiv.tex
\documentclass[runningheads]{llncs}

\usepackage[mobile]{eccv}

\usepackage{eccvabbrv}

\usepackage{graphicx}
\usepackage{booktabs}

\usepackage[accsupp]{axessibility}  

\usepackage{hyperref}

\usepackage{orcidlink}

\input{preamble}

\begin{document}

\title{TIMAR: Causal Turn-Level Modeling of Interactive 3D  Conversational Head Dynamics} 

\titlerunning{Towards Seamless Interaction}

\author{
Junjie Chen\inst{1,2}\orcidlink{0009-0001-5288-048X} \and
Fei Wang\inst{1,2} \and
Zhihao Huang\inst{5,6} \and
Qing Zhou\inst{8} \and
Kun Li\inst{7} \and
Dan Guo\inst{1} \and
Linfeng Zhang\inst{4} \and
Xun Yang\inst{3}
}

\authorrunning{J.~Chen et al.}

\institute{
$^1$HFUT\quad 
$^2$IAI, Hefei Comprehensive National Science Center\quad
$^3$USTC \quad
$^4$SJTU \quad  \\
$^5$TeleAI, China Telecom \quad
$^6$NWPU \quad
$^7$UAEU \quad
$^8$AHPU \quad 
}

\maketitle

\input{sec/0_arxiv_abs}
\input{sec/1_intro}
\input{sec/2_related_work}

\input{sec/3_method}

\input{sec/4_exp}
\input{sec/5_conclusion}

%
%
\bibliographystyle{splncs04}
\bibliography{main}

\input{sec/X_suppl}
\end{document}

%% file: preamble.tex
\usepackage{lipsum}
\usepackage[table]{xcolor}
\usepackage{ifthen}
\usepackage{calc}
\usepackage{siunitx}

\makeatletter
\newcommand{\absnum}[1]{%
  \begingroup
    \dimen0=\dimexpr#1pt\relax
    \ifdim\dimen0<0pt
      \dimen0=-\dimen0
    \fi
    \edef\abs@temp{\strip@pt\dimen0}%
    \num{\abs@temp}%
  \endgroup
}
\newcommand{\updiff}[1]{%
  \ifthenelse{\lengthtest{#1 pt > 0 pt}}{%
    \textsuperscript{\colorbox{green!10}{\scriptsize$\uparrow$\,\absnum{#1}}}%
  }{%
    \textsuperscript{\colorbox{red!10}{\scriptsize$\downarrow$\,\absnum{#1}}}%
  }%
}
\newcommand{\downdiff}[1]{%
  \ifthenelse{\lengthtest{#1 pt < 0 pt}}{%
    \textsuperscript{\colorbox{green!10}{\scriptsize$\downarrow$\,\absnum{#1}}}%
  }{%
    \textsuperscript{\colorbox{red!10}{\scriptsize$\uparrow$\,\absnum{#1}}}%
  }%
}
\makeatother

\newcommand{\headcell}[2]{
  \parbox[c][2.6em][c]{\widthof{\textbf{#1}}}{
    \centering \textbf{#1}\\[-0.1em]\small #2
  }
}

\usepackage{wrapfig}

%% file: sec/0_arxiv_abs.tex
\begin{abstract}
Human conversation involves continuous exchanges of speech and nonverbal cues such as head nods, gaze shifts, and facial expressions that convey attention and emotion.  
Modeling these bidirectional dynamics in 3D is essential for building expressive avatars and interactive robots.  
However, existing frameworks often treat talking and listening as independent processes or rely on non-causal full-sequence modeling, hindering temporal coherence across turns.  
We present \textbf{TIMAR} (\textbf{T}urn-level \textbf{I}nterleaved \textbf{M}asked \textbf{A}uto\textbf{R}egression), a causal framework for 3D conversational head generation that models dialogue as interleaved audio-visual contexts.  
It fuses multimodal information within each turn and applies turn-level causal attention to accumulate conversational history, while a lightweight diffusion head predicts continuous 3D head dynamics that captures both coordination and expressive variability.  
Experiments on the DualTalk benchmark show that TIMAR achieves 15-30\% relative improvements on the test set and maintains comparable gains on out-of-distribution data.
\textit{The source code has been released at \href{https://github.com/CoderChen01/towards-seamless-interaction}{CoderChen01/towards-seamless-interaction}.}
\keywords{3D conversational head generation \and Speech-driven facial motion synthesis \and Dual-speaker interaction}
\end{abstract}

%% file: sec/1_intro.tex
\input{figs/teaser}
\section{Introduction}
\label{sec:intro}
Human conversation is an intricate interplay of speech and facial behavior.  
Beyond verbal communication, subtle nonverbal signals such as head nods, gaze shifts, and micro-expressions continuously convey intent, attention, and empathy~\cite{burgoon2021nonverbal}.  
Modeling these bidirectional dynamics is essential for embodied conversational agents, social robots, and immersive telepresence systems that must \emph{listen, react, and respond} naturally in a streaming conversational setting~\cite{skantze2021turn}.

Recent advances in 3D talking-head generation~\cite{fan2022faceformer,xing2023codetalker,li2025towards,sun2024diffposetalk,ye2024realdportrait,chen2025cafetalk} and listening-head synthesis~\cite{zhou2022responsive,liu2023mfr,ng2022learning,siniukov2025ditailistener} have significantly improved visual realism and speech synchronization.  
However, most frameworks still treat these two processes, talking and listening, as independent directions of motion generation, lacking a unified temporal model that captures their mutual influence.  
As illustrated in \Cref{fig:teaser}, talking-head systems generate motion only from a speaker's own audio, and listening-head systems react only to the interlocutor, while natural conversation emerges from their intertwined evolution.  
Even the recent DualTalk~\cite{peng2025dualtalk} framework, though jointly modeling both speakers, relies on bidirectional attention over full conversations.
Such a formulation is effective for offline synthesis but less suited for causal or streaming generation, where models must respond turn by turn to ongoing dialogue.

Our core motivation is that conversational behavior unfolds through causally linked \emph{turns}~\cite{skantze2021turn},
where each turn's facial motion depends on both speakers' preceding speech and visual cues,
reflecting how humans naturally coordinate responses through continuous multimodal feedback. 
To reflect this principle, we formulate \emph{interactive 3D conversational head generation}\footnote{A problem statement is provided in Appendix~\cref{app:problem-statement} for clarity.} as a turn-level causal process, aligning computational modeling with the temporal logic of human interaction.
As shown in \Cref{fig:teaser}, we introduce \textbf{TIMAR}, an \emph{autoregressive-diffusion} framework that couples \emph{masked}, \emph{turn-level causal} modeling over \emph{interleaved} audio-visual tokens with diffusion-based decoding of continuous 3D agent head.
TIMAR represents a conversation as an interleaved sequence of multimodal tokens from both participants, segmented at the turn level.  
The model fuses intra-turn audio-visual information bidirectionally while maintaining causal dependencies across turns,  
and predicts the agent's 3D head using a lightweight diffusion-based generative head conditioned on the fused context.  
This formulation enables the model to accumulate conversational history and reason over conversational flow.
Our approach introduces three main contributions:
\begin{itemize}
\item \textbf{Turn-level causal formulation.}  
We formulate interactive 3D head generation as a causal, turn-wise prediction problem,  
enforcing strict temporal consistency and supporting streaming-compatible generation.
\item \textbf{Interleaved multimodal fusion.}  
We design an interleaved audio-visual context that encodes both speakers' speech and 3D head tokens,  
enabling the model to learn \emph{intra-turn alignment} and \emph{inter-turn dependency} under causal constraints for coherent conversational modeling.
\item \textbf{Lightweight diffusion-based generative decoding.}  
We introduce a compact diffusion-based decoder that models 3D head motion as a continuous probabilistic process, capturing natural variability while maintaining temporal coherence across conversational turns.
\end{itemize}

Compared with DualTalk~\cite{peng2025dualtalk}, which processes entire dialogues with full-sequence modeling, TIMAR's causal formulation enables natural and streaming-capable generation that mirrors real conversational timing and feedback.
Extensive evaluations demonstrate consistent improvements in realism, synchronization, and responsiveness across both in-distribution and unseen scenarios.

%% file: figs/teaser.tex
\begin{figure}
\centering
\includegraphics[width=\textwidth]{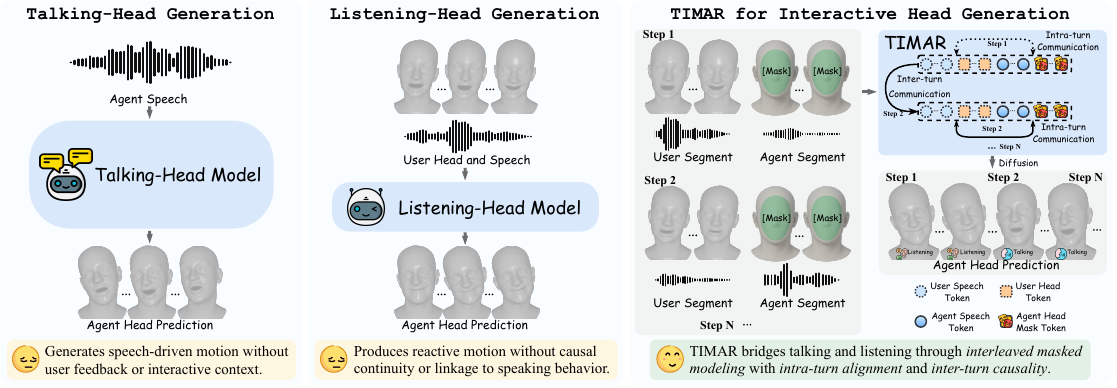}
\caption{
\textbf{Comparison of head generation paradigms and our TIMAR framework.}  
Prior paradigms treat talking and listening as separate processes:  
\emph{Talking-Head Generation} produces speech-driven motion without user feedback,  
while \emph{Listening-Head Generation} yields reactive behavior without causal continuity.  
\textbf{TIMAR} unifies both within an \textit{interleaved masked} and \textit{causally grounded} framework,  
modeling conversation as sequential turns of interleaved user-agent audio-visual tokens.  
It achieves intra-turn alignment through bidirectional fusion and inter-turn dependency through causal attention,  
producing coherent and contextually responsive 3D head motion.
}
\label{fig:teaser}
\end{figure}

%% file: sec/2_related_work.tex
\section{Related Work}
\label{sec:related_work}

\subsection{3D Talking- and Listening-Head Generation}

\subsubsection{Talking-Head Generation.}
Early works on \emph{talking-head generation} aim to synthesize a speaker's facial motion from speech, producing temporally aligned and expressive visual outputs~\cite{zhou2020makelttalk,prajwal2020lip,zhou2021pose}.
A large body of research operates directly in the RGB domain, learning mappings from audio to lip movements and facial expressions in videos~\cite{zhou2020makelttalk,prajwal2020lip,zhou2021pose,he2024gaia,xu2024hallo,xu2024vasa,chen2025echomimic,wang2025omnitalker,cheng2025dawn}.
Representative methods such as VASA-1~\cite{xu2024vasa}, Hallo~\cite{xu2024hallo}, and EchoMimic~\cite{chen2025echomimic} have demonstrated realistic speech-driven head animations.
To enhance controllability and geometric consistency, several studies~\cite{huang2022audio-driven,zhang2023sadtalker,chen2024emotion-aware,wei2024aniportrait} introduce 3D Morphable Models (3DMM)~\cite{blanz2023morphable,egger20203d,blanz2003face} as intermediate representations, predicting 3DMM parameters conditioned on audio signals.
More recent works advocate for direct generation in 3D space, which enables physically grounded motion synthesis that can be directly applied to robotic heads, virtual avatars, and psychological or affective behavior analysis~\cite{niswar2009real,karras2017audio,cudeiro2019capture,richard2021meshtalk,fan2022faceformer,ye2023geneface,danvevcek2023emotional,peng2023selftalk,thambiraja2023imitator,stan2023facediffuser,xing2023codetalker,peng2023emotalk,li2023one,ye2024realdportrait,aneja2024facetalk,yang2024probabilistic,sungbin24_interspeech,sun2024diffposetalk,zhou2024meta,sung2024laughtalk,peng2024synctalk,li2025towards,li2025instag,xie2025ecoface,chen2025cafetalk,chae2025perceptually}.
For instance, FaceFormer~\cite{fan2022faceformer}, CodeTalker~\cite{xing2023codetalker}, DiffPoseTalk~\cite{sun2024diffposetalk}, and TexTalker~\cite{li2025towards} adopt transformer- or diffusion-based frameworks~\cite{vaswani2017attention,ho2020denoising} to learn continuous 3D head dynamics from speech.

\subsubsection{Listening-Head Generation.}
Parallel to talking-head research, another line of work explores \emph{listening-head generation}, which models non-verbal feedback of a listener conditioned on the interlocutor's speech or facial motion.
These systems aim to reproduce subtle and socially meaningful behaviors such as nodding, gaze shifts, and micro-expressions that convey attention, agreement, or emotional alignment~\cite{sonlu2021conversational,bohus2010facilitating,cassell1994animated,liu2024customlistener,huang2022perceptual,liu2023mfr,ng2022learning,song2023emotional,siniukov2025ditailistener,zhou2022responsive}.
Early approaches learn reactive behaviors directly from 2D videos~\cite{sonlu2021conversational,bohus2010facilitating,cassell1994animated}, while recent studies employ 3DMM- or mesh-based representations to achieve more geometrically consistent listener motion~\cite{zhou2022responsive,ng2022learning,liu2023mfr,tran2024dim,wang2025diffusion}.

Despite these advances, most talking- and listening-heads are treated as \textit{separate processes}, with one responsible for speaking and the other for responding, rather than capturing the intertwined temporal dynamics that define genuine dyadic communication.
In contrast, our work focuses on unified modeling of interactive 3D head dynamics under an interleaved conversational structure.

\subsection{Modeling Interactive 3D Conversational Heads}

Modeling conversational interaction is essential for producing coherent and socially responsive 3D head dynamics~\cite{agrawal2025seamless,zhou2025interactive}.
Human dialogue is inherently bidirectional and temporally dependent, yet most generative frameworks still treat each participant independently, failing to capture the continuous exchange of verbal and non-verbal cues~\cite{sacks1974simplest,clark1991grounding,stivers2009universals,vaswani2017attention,ho2020denoising,skantze2021turn}.
Some 2D-based studies (\textit{e.g.}, INFP~\cite{zhu2025infp}, ARIG~\cite{guo2025arig}) have explored mutual head or gesture coordination between interlocutors, but these methods remain limited to image-space motion and lack explicit 3D geometry control.
In 3D domains, DualTalk~\cite{peng2025dualtalk} represents an important step toward dual-speaker modeling, integrating both speaking and listening behaviors within a unified framework.
However, its bidirectional full-sequence processing is designed for offline synthesis and is less suited for causal or streaming generation.
Our framework models conversations as causally conditioned turns, enabling temporally coherent and streaming-capable 3D head generation through interleaved autoregressive diffusion.

%% file: sec/3_method.tex
\section{The TIMAR Framework}

As shown in \Cref{fig:timar}, TIMAR discretizes speech and encodes 3D head parameters into a shared token space, segments them into fixed-length turns, and interleaves user-agent streams. 
A turn-level fusion module models \emph{intra-turn alignment} and \emph{inter-turn dependency} under causal masking, while a diffusion head denoises the masked agent head from the fused context. 

\subsection{Interleaved Audio-Visual Context}

Given a $T$-second conversational segment sampled at an audio rate $f_{s}$ and a motion frame rate $f_{h}$, 
let the user's speech and 3D head motion be $S^{u} \in \mathbb{R}^{T f_{s}}$ and $H^{u} \in \mathbb{R}^{T f_{h}\times d_{h}}$,
and the agent's speech and motion be $S^{a} \in \mathbb{R}^{T f_{s}}$ and $H^{a} \in \mathbb{R}^{T f_{h}\times d_{h}}$,
where $d_{h}$ denotes the dimensionality of the 3D head representation.
TIMAR first aligns all modalities in a shared token space using a pretrained \textit{speech tokenizer} and a learnable \textit{3D head motion encoder}, 
then constructs an Interleaved Audio-Visual Context that provides the turn-level multimodal input for autoregressive generation.

\subsubsection{Speech Tokenizer.}
We define a speech tokenizer $\mathcal{F}_{\text{speech}}$ built on a pretrained model $\mathcal{M}_{\text{speech}}$ with token dimension $d_{s}$.  
To align the extracted speech features with the shared $d_{t}$-dimensional token space, a learnable projection $\mathcal{P}_{\text{speech}}:\mathbb{R}^{d_{s}}\!\rightarrow\!\mathbb{R}^{d_{t}}$ is applied after temporal alignment. 
Specifically, $\mathrm{interp}_{f_{h}}(\cdot)$ temporally resamples the speech features to match the motion frame rate $f_{h}$, ensuring synchronized alignment across modalities. 
Given a speech $S$, the $\mathcal{F}_{\text{speech}}$ produces the token sequence as
\begin{equation}
\mathbf{S} = \mathcal{P}_{\text{speech}}\!\left(\mathrm{interp}_{f_{h}}\!\big(\mathcal{M}_{\text{speech}}(S)\big)\right),
\quad \mathbf{S}\in\mathbb{R}^{T f_{h}\times d_{t}}.
\end{equation}

We denote the resulting speech token sequences for the user and the agent as $\mathbf{S}^{u}$ and $\mathbf{S}^{a}$ respectively.

\subsubsection{3D Head Motion Encoder.}
To embed 3D head motion into the same token space,  
we introduce an encoder $\mathcal{F}_{\text{head}}$ that maps the 3D head parameters of each frame to a $d_{t}$-dimensional token representation:
\begin{equation}
\mathbf{H} = \mathcal{F}_{\text{head}}(H),
\quad \mathbf{H}\in\mathbb{R}^{T f_{h}\times d_{t}}.
\end{equation}

We denote the corresponding user and agent 3D head motion token sequences as $\mathbf{H}^{u}$ and $\mathbf{H}^{a}$, respectively.  

\subsubsection{Interleaving.}
We segment the audio-visual context sequences into $N=T/c$ chunks of $c$ seconds each\footnote{We preprocess data such that $T$ is divisible by $c$.}.
For the $i$-th chunk, we define:
\begin{equation}
\begin{aligned}
\mathcal{S}^{u}_{i} &= \mathbf{S}^{u}_{[(i-1)cf_{h}:icf_{h}]}, \quad
\mathcal{S}^{a}_{i} = \mathbf{S}^{a}_{[(i-1)cf_{h}:icf_{h}]}, \\
\mathcal{H}^{u}_{i} &= \mathbf{H}^{u}_{[(i-1)cf_{h}:icf_{h}]}, \quad
\mathcal{H}^{a}_{i} = \mathbf{H}^{a}_{[(i-1)cf_{h}:icf_{h}]}.
\end{aligned}
\end{equation}

In practice, each chunk is encoded independently rather than from the full sequence,
ensuring that no future information is exposed during tokenization.

Finally, the interleaving function $\mathcal{F}_{\text{interleave}}$ constructs the interleaved audio-visual context token sequence $\mathcal{T}$ as:
\begin{equation}
\begin{split}
\mathcal{T}
 &= \mathcal{F}_{\text{interleave}}(\mathcal{S}^{u},\mathcal{S}^{a},\mathcal{H}^{u},\mathcal{H}^{a})
 = \left(\mathbf{T}_{i}\right)_{i=1}^{N}, \\
&\text{where} \quad
\mathbf{T}_{i} = \left(\mathcal{S}^{u}_{i}, \mathcal{S}^{a}_{i}, \mathcal{H}^{u}_{i}, \mathcal{H}^{a}_{i}\right).
\end{split}
\end{equation}

The resulting $\mathcal{T}$ provides temporally aligned, turn-level interleaved audio-visual tokens across both participants, 
forming the multimodal conversational context that drives the causal autoregressive generation in TIMAR.

\input{figs/timar-framework}

\subsection{Turn-Level Causal Multimodal Fusion}
Given the token sequence $\mathcal{T}$, 
the Turn-Level Causal Multimodal Fusion module, denoted as $\mathcal{F}_{\text{fusion}}$, 
is designed to perform intra-turn alignment and capture inter-turn dependencies under turn-level causal constraints.

\input{figs/tlca}

\subsubsection{Positional Embedding.}
To encode both intra-turn and inter-turn positional relations, 
we introduce a learnable positional embedding $P_{1}$ that provides explicit temporal awareness for each token in $\mathcal{T}$. 
This positional embedding allows the model to distinguish not only the token order within a turn but also the relative positions across turns, 
facilitating temporally consistent contextual reasoning.

\subsubsection{Turn-Level Causal Attention.}
The fusion process is implemented by a stacked Transformer encoder $\mathcal{E}$ equipped with our proposed \textit{Turn-Level Causal Attention (TLCA)}. 
As illustrated in \Cref{fig:tlca}, TLCA enables bidirectional attention among tokens within the same turn to achieve fine-grained speech-motion alignment, 
while constraining attention across turns to be strictly causal, ensuring that each turn can only attend to preceding ones. 
This design allows the encoder to learn short-term multimodal synchronization and long-term conversational dependency jointly.

\subsubsection{Fusion Process.}
Omitting normalization and residual connections for brevity, 
the $\mathcal{F}_{\text{fusion}}$ can be expressed as:
\begin{equation}
\mathcal{Z}
= \mathcal{F}_{\text{fusion}}(\mathcal{T})
= \mathcal{E}(\mathcal{T}+P_{1})
= \left(\mathbf{Z}_{i}\right)_{i=1}^{N},
\end{equation}
where $\mathbf{Z}_{i}$ denotes the fused representation of the $i$-th turn, 
and $\mathcal{Z}$ represents the temporally integrated multimodal feature sequence. 
These features serve as the bottleneck representations that condition the diffusion head to model the per-token probability distribution of 3D agent head.

\subsection{Lightweight Diffusion Head}
As illustrated in \Cref{fig:timar}~(left), we introduce the Lightweight Diffusion Head $\mathcal{F}_{\text{diff}}$, 
which models masked agent head through conditional denoising in a continuous parameter space.  
During training, we randomly select a subset of agent head positions and replace each selected position with the same learnable \emph{mask token} $\mathbf{h}^{m}$.  
For any masked position $i$, the fused contextual representation produced by $\mathcal{F}_{\text{fusion}}$ is denoted as $\mathbf{z}^{m}_{i}$,  
and the corresponding ground-truth agent head parameter is written as $\mathbf{h}^{a}_{i}$.  
Inspired by Li \etal~\cite{li2024autoregressive}, we model the conditional distribution  
$p(\mathbf{h}^{a}_{i}\mid \mathbf{z}^{m}_{i})$ through a \emph{diffusion process} rather than \emph{regression},  
allowing the model to capture the intrinsic stochasticity and multimodality of natural 3D facial motion without relying on discrete quantization.

Given $\mathbf{z}^{m}_{i}$, $\mathcal{F}_{\text{diff}}$ predicts the clean agent head parameters at position $i$ via conditional denoising.  
To enable $\mathcal{F}_{\text{diff}}$ to be aware of the frame index of each masked position,  
we introduce a learnable positional embedding $P_{2}$ and add the corresponding vector $P_{2}^{(i)}$ to the conditioning feature.  
The per-frame prediction process is then formulated as
\begin{equation}
\hat{\mathbf{h}}^{a}_{i}
= \mathcal{F}_{\text{diff}}(\mathbf{z}^{m}_{i}, \tau)
= \epsilon_{\theta}(x_{\tau}\mid \tau, \mathbf{z}^{m}_{i} + P_2^{(i)}),
\end{equation}
where $\tau$ is the diffusion timestep and $x_{\tau}$ is a noisy version of the ground-truth parameter $\mathbf{h}^{a}_{i}$ produced by a predefined forward noise schedule.  
During sampling, $x_{\tau}$ is initialized from pure noise and iteratively denoised to recover the final 3D head parameters.  
Implemented as a lightweight MLP, $\epsilon_{\theta}$ performs efficient token-wise diffusion in a continuous parameter space,  
bridging multimodal conversational context and geometric reconstruction to generate stochastic yet temporally coherent 3D head dynamics.

\subsection{TIMAR Training Recipe}
We now describe the training strategy of TIMAR for learning 3D conversational dynamics via multimodal fusion and lightweight diffusion-based reconstruction. 
The model is optimized to recover masked agent head parameters conditioned on interleaved conversational context. 
Full details are provided in Appendix~\cref{app:timar_details}.

\subsubsection{Masking Strategy.}
During training, a fixed proportion $r$ of the agent head tokens is randomly replaced by a learnable \textit{mask token} $\mathbf{h}^{m}$. 
This random masking encourages the model to learn robust token-wise completion and to generalize across different conversational contexts. 
The masked tokens are processed through the multimodal fusion and diffusion modules, 
enabling the model to reconstruct the corresponding ground-truth 3D agent head parameters conditioned on the visible conversational context.

\subsubsection{Optimization Objective.}
We minimize a \emph{diffusion objective} under the $x_{0}$-prediction parameterization, operating directly on the continuous 3D head sequences.  
We use $f_{\theta}$ to denote the end-to-end process illustrated in \Cref{fig:timar}~(left), which includes speech and head encoding, interleaving, masking, multimodal fusion, and lightweight diffusion-based reconstruction.  
Given the conversational inputs $(S^{u}, S^{a}, H^{u}, H^{a})$, let $\mathcal{K}$ denote the index set of masked agent-head positions.  
For each $i \!\in\! \mathcal{K}$, the clean target is the ground-truth agent head parameter $H^{a}_{i}$, which is perturbed within $\mathcal{F}_{\text{diff}}$ through the forward diffusion process $q(x_{\tau_i} \mid H^{a}_{i})$ using a randomly sampled timestep $\tau_{i}$.  
Conditioned on the raw multimodal inputs and $\tau_{i}$, the model predicts the denoised estimate at position $i$, yielding the per-sample diffusion loss:
\begin{equation}
\label{eq:timar_x0_obj}
\mathcal{L}_{\text{diff}}
=
\frac{1}{|\mathcal{K}|}
\sum_{i\in\mathcal{K}}
\mathbb{E}_{\tau_{i}}
\left[
\big\|
H^{a}_{i}
-
f_{\theta}^{(i)}(S^{u}, S^{a}, H^{u}, H^{a}, \tau_{i})
\big\|_{2}^{2}
\right].
\end{equation}
Under the learned-variance setting, the final objective also includes a variational bound term, yielding a combined loss on mean and variance predictions.

\subsubsection{Classifier-Free Guidance~(CFG).}
With a fixed probability $p_{\text{cfg}}$, the user's entire set of tokens is replaced by a shared learnable \textit{fake token} $\mathbf{h}^{f}$ during training. 
This stochastic substitution constructs an unconditional training branch that removes contextual cues from the user side, 
allowing the model to learn both context-dependent and context-independent 3D agent head generation. 

\subsection{TIMAR Sampling}

As shown in~\Cref{fig:timar}~(right), during sampling, TIMAR performs turn-wise autoregressive generation using two components:
a \textit{conversational stream} that provides the incoming multimodal inputs,
and a \textit{context token buffer} that stores previously processed tokens for temporal conditioning.

\subsubsection{Turn Construction.}
At each conversational turn $t$, we collect a $c$-second segment of user speech $S^{u}_{t}$, agent speech $S^{a}_{t}$, and user head motion $H^{u}_{t}$.  
These signals are processed by the speech tokenizer $\mathcal{F}_{\text{speech}}$ and head encoder $\mathcal{F}_{\text{head}}$ to obtain the corresponding token sequences 
$\mathbf{S}^{u}_{t}$, $\mathbf{S}^{a}_{t}$, and $\mathbf{H}^{u}_{t}$.  
The agent heads of this turn is filled with a learnable mask token sequence 
$\mathbf{H}^{m}_{t}=\left(\mathbf{h}^{m}_{t,i}\right)_{i=1}^{K}$, where $K=c f_{h}$ denotes the number of frames in the turn.  
The current turn is represented as 
$\mathbf{T}_{t}=\left(\mathbf{S}^{u}_{t},\mathbf{S}^{a}_{t},\mathbf{H}^{u}_{t},\mathbf{H}^{m}_{t}\right)$.  
To provide historical context, a \textit{context token buffer} stores the tokenized turns from the previous $n$ steps, 
$\left(\mathbf{T}_{t-n},\dots,\mathbf{T}_{t-1}\right)$, where $n$ is a hyperparameter controlling the context history length.  
The input for the current turn is then constructed as 
\begin{equation}
\mathcal{T}_{t}=\left(\mathbf{T}_{t-n},\dots,\mathbf{T}_{t-1},\mathbf{T}_{t}\right).
\end{equation}
For all previous turns in the buffer, the agent head tokens remain filled with mask tokens rather than the predicted ones.  
This prevents the accumulation of autoregressive errors and ensures that the model relies solely on reliable conversational context rather than potentially compounding errors.

\subsubsection{Turn-Level Autoregressive Diffusion Sampling.}
At each conversational turn $t$, the interleaved token sequence $\mathcal{T}_{t}$ is first processed by the fusion module $\mathcal{F}_{\text{fusion}}$ 
to obtain the fused representation $\mathbf{Z}_{t}$. 
The features corresponding to the masked agent-head tokens, denoted as $\mathbf{Z}^{m}_{t}$, 
are then fed into the diffusion head $\mathcal{F}_{\text{diff}}$, which performs iterative denoising across diffusion timesteps. 
Starting from Gaussian noise, the diffusion process progressively refines the latent variables conditioned on $\mathbf{Z}^{m}_{t}$, 
recovering the predicted 3D agent head parameters $\hat{H}^{a}_{t}$ for the current turn. 

\subsubsection{Sampling with CFG.}
The CFG-based sampling adjusts the strength of contextual conditioning during iterative denoising.  
For each conversational turn $t$, the unconditional features $\overline{\mathbf{Z}}^{m}_{t}$ are obtained by replacing all user tokens in $\mathcal{T}_{t}$ 
with a fake token $\mathbf{h}^{f}$.  
At each diffusion step $\tau$, let $X_{\tau}$ denote the current noisy estimate of the 3D head parameters.  
The denoising update is formulated as
\begin{equation}
\epsilon_{\theta}(X_{\tau}\!\mid\!\tau,\overline{\mathbf{Z}}^{m}_{t})
+ \omega \big[
\epsilon_{\theta}(X_{\tau}\!\mid\!\tau,\mathbf{Z}^{m}_{t})
- \epsilon_{\theta}(X_{\tau}\!\mid\!\tau,\overline{\mathbf{Z}}^{m}_{t})
\big],
\end{equation}
where $\omega$ is the guidance scale that controls the trade-off between contextual adherence and generative diversity.
For brevity, the addition of positional embeddings to the conditioning features is omitted in the above expression.

%% file: figs/timar-framework.tex
\begin{figure*}[t]
\centering
\includegraphics[width=\linewidth]{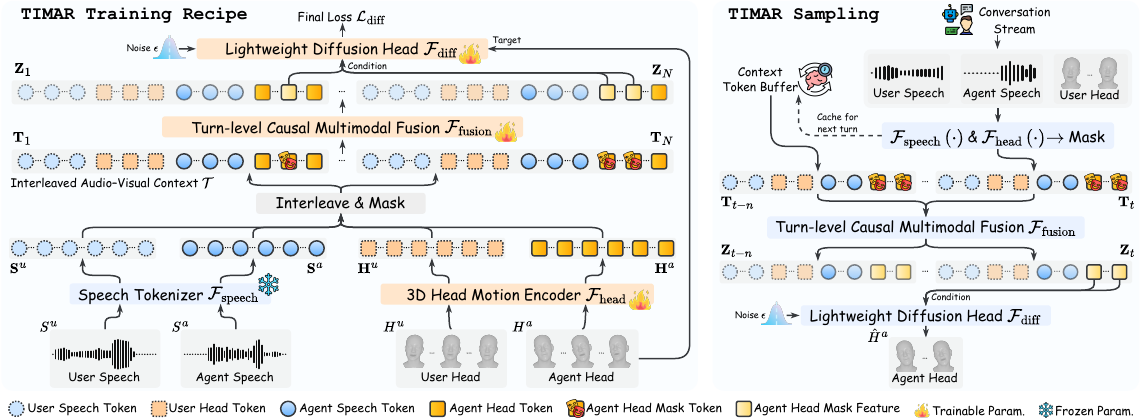}
\caption{
\textbf{The architecture and workflow of TIMAR.}
TIMAR models interactive 3D conversational head dynamics through a \emph{turn-level, causal, and interleaved} generation process. 
In training (\textbf{left}), the speech and head motions of both user and agent are encoded into a shared token space, interleaved by conversational turns, with the agent head tokens masked.
The \textit{Turn-level Causal Multimodal Fusion} module fuses audio-visual context bidirectionally within each turn and causally across turns, producing masked-agent features that condition the \textit{Lightweight Diffusion Head} to learn the head motion distribution. 
In sampling (\textbf{right}), the model caches history tokens and autoregressively denoises each new turn, yielding temporally coherent and context-aware 3D head motion generation in streaming conversation.
}
\label{fig:timar}
\end{figure*}

%% file: figs/tlca.tex
\begin{wrapfigure}{l}{0.43\textwidth}
\centering
\vspace{-1em}
\includegraphics[width=0.88\linewidth]{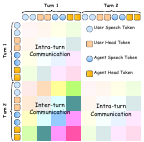}
\caption{
\textbf{Illustration of Turn-Level Causal Attention (TLCA).}
The example shows two turns with two tokens per modality (user speech, user head, agent speech, and agent head). 
Different color blocks represent modality-wise token communication. 
TLCA models both \textit{intra-turn communication} through bidirectional attention and \textit{inter-turn communication} through turn-level causal attention to capture temporal dependencies without future leakage.
}
\label{fig:tlca}
\vspace{-5em}
\end{wrapfigure}

%% file: sec/4_exp.tex
\section{Experiments}
\label{sec:exp}
\subsubsection{TIMAR Default Configuration.}
Unless otherwise specified, all experiments are conducted under a unified default setup.  
$\mathcal{M}_{\text{speech}}$ uses wav2vec 2.0~\cite{baevski2020wav2vec}. 
The shared token dimension is set to $d_{t}=1024$.  
Speech and motion sequences are sampled at $f_{s}=16$~kHz and $f_{h}=25$~fps, respectively.  
Each training conversation sample spans $T=8$~s and is divided into $N=8$ fixed-length turns with chunk duration $c=1$~s.  
Each turn contains temporally aligned user and agent audio-visual segments forming one interleaved context unit for training.  
During sampling, the context history expands progressively over $n$ previous turns (up to $n=7$).
Implementation details are provided in Appendix~\cref{app:timar_details}.

\subsubsection{Benchmark Setup.}
We follow the experimental setup of the interactive 3D conversational head generation benchmark proposed by DualTalk~\cite{peng2025dualtalk}.  
All models are trained on their official training split and evaluated on the provided \emph{test dataset} and an additional \emph{out-of-distribution~(OOD) dataset} to assess generalization.  
The 3D head representation is represented by 56 FLAME parameters~\cite{li2017learning}, including 50 expression, 3 jaw, and 3 head pose dimensions per frame.
Dataset details are provided in Appendix~\cref{app:dataset_details}.

\subsubsection{Evaluation Metrics.}
We use the same evaluation metrics as DualTalk, including Fr\'{e}chet Distance (FD), Paired Fr\'{e}chet Distance (P-FD),  
Mean Squared Error (MSE), SI for Diversity (SID), and Residual Pearson Correlation Coefficient (rPCC).  
All metrics are computed separately for the expression~(EXP), jaw~(JAW), and head pose~(POSE) components of the FLAME parameters.  
Together, these metrics assess the realism, temporal synchronization, motion diversity, and expression accuracy of generated 3D head dynamics.  
Metric details are provided in Appendix~\cref{app:metric_details}.

\input{tabs/main-results}

\subsubsection{Comparison Protocol.}
We compare with FaceFormer~\cite{fan2022faceformer}, 
CodeTalker~\cite{xing2023codetalker}, 
EmoTalk~\cite{peng2023emotalk}, 
SelfTalk~\cite{peng2023selftalk}, 
L2L~\cite{ng2022learning}, 
and DualTalk~\cite{peng2025dualtalk} under the official DualTalk benchmark metrics on both the \emph{test} and \emph{OOD} datasets. 
For fair comparison with DualTalk, the only prior state-of-the-art dual-speaker interactive model, we adopt a \emph{progressive-context streaming evaluation}, 
where dialogue history is accumulated up to a fixed limit ($n=7$) under causal inference. 
This protocol is applied only to DualTalk because the other baselines are single-speaker talking or listening models 
that do not model cross-turn interaction, and adapting them to streaming would not change their conditioning. 
All methods are evaluated on temporally aligned sequences; 
for turn-based generation, remaining frames are padded with the final predicted 3D head parameters.

\subsubsection{TIMAR achieves consistent gains on both streaming and full-sequence benchmarks.}
As shown in \Cref{tab:main-results}, under progressive-context streaming inference ($n=0,3,7$), TIMAR consistently outperforms DualTalk on both \emph{test} and \emph{OOD} datasets, yielding about 15--30\% lower FD/P-FD on \emph{test} and around 5--10\% lower FD/P-FD on \emph{OOD}. 
In \Cref{tab:other-baselines}, TIMAR achieves the best overall results among all compared baselines on both datasets, with the lowest FD/P-FD and rPCC while maintaining competitive or higher SID. 
Overall, these results show that turn-level causal modeling improves realism and inter-speaker synchronization, and generalizes better to unseen conversations.

\input{tabs/other-baselines}
\input{figs/qulitative_comp_context}

\subsubsection{TIMAR yields more coherent interaction and is consistently preferred by users.}
As shown in \Cref{fig:qulitative_comp_context}, under progressive-context streaming ($n{=}0,3,7$), TIMAR produces more context-consistent behaviors than DualTalk, with more stable and context-aligned head dynamics as history increases. 
For perceptual evaluation, we randomly sampled 25 clips from the \emph{test} dataset and 25 from the \emph{OOD} dataset, and recruited 10 participants, resulting in 500 paired A/B evaluations. 
Participants rated both methods on a 1–5 scale across Motion Naturalness, Facial Expression Naturalness, Interaction Naturalness, and Lip-sync Accuracy. 
Preference rates (ties counted as 0.5) with 95\% bootstrap confidence intervals show consistent preference for TIMAR across all criteria. 
Implementation details of the user study are provided in Appendix~\cref{app:user-study}.

\input{figs/cfg_case}

\subsubsection{Impact of classifier-free guidance.}
As shown in \Cref{fig:cfg}, increasing properly $\omega$ improves both FD and P-FD metrics, 
reflecting stronger contextual adherence and better alignment between the agent's responses and the user's multimodal inputs.  
Starting from $\omega{=}0$, which corresponds to unconditional sampling, the generated heads appear less responsive and contextually ambiguous.  
As $\omega$ increases properly, the agent exhibits progressively richer expressions and more synchronized motion with the interlocutor,  
demonstrating the benefit of conditional modulation on interactive dynamics.  
In contrast, DualTalk lacks a CFG mechanism and thus cannot adjust its contextual conditioning during inference,  
resulting in fixed, non-controllable generation.  

\input{figs/param_perf--ablation-diff-mlp-loss}

\subsubsection{Performance comparison across parameter scales.}
To verify that our performance gains stem from improved modeling rather than increased capacity,
we compare TIMAR and DualTalk across parameter scales, as shown in \Cref{fig:param-perf}.
Results show that enlarging DualTalk does not yield consistent improvements in FD or P-FD,
while TIMAR achieves lower errors under comparable parameter counts.
These findings indicate that the proposed causal formulation and interleaved fusion enable more effective learning without relying on model size.

\input{tabs/main-ablation}

\subsubsection{Ablation studies on core design choices.}
The results are summarized in \Cref{tab:main-ablation,fig:ablation-diffusion-head-mlp-head-loss}.
Replacing the diffusion head with a direct MLP predictor results in smoother training loss but inferior test performance, indicating overfitting and weaker generalization. 
The diffusion-based formulation achieves better FD and P-FD scores by capturing the stochastic nature of conversational motion.  
When substituting the proposed Turn-Level Causal Attention (TLCA) with full bidirectional attention, temporal consistency slightly deteriorates, demonstrating the necessity of causal masking for progressive generation.
Adopting an asymmetric encoder–decoder design following MAE~\cite{he2021masked} does not improve results, and the encoder-only configuration yields better performance.
This comparison is motivated by the masked-prediction paradigm of MAE, allowing us to test whether an additional decoder benefits temporal reasoning.

\input{figs/robustness}

\subsubsection{Robustness and streaming latency.}
We evaluate robustness on the \emph{test} dataset by perturbing user head tokens to simulate tracking failures (\Cref{fig:head_motion_failure}). 
For random frame dropping, we replace missing frames with the last observed frame; under this setting, degradation is mild for both methods due to temporal smoothness, while TIMAR consistently shows smaller P-FD increase. 
With injected noise, DualTalk deteriorates rapidly as corruption increases, whereas TIMAR maintains substantially lower degradation across noise levels. 
Under speech failures (\Cref{fig:speech_failure}), silencing user speech has limited impact on both models. 
In contrast, silencing agent speech leads to catastrophic degradation in DualTalk because its generation heavily conditions on the agent's own speech features; removing this signal breaks its primary driving input. 
TIMAR exhibits much smaller performance drop, as its turn-level causal modeling leverages conversational context beyond a single speech stream. 
Regarding streaming, TIMAR synthesizes one second of motion (25 frames) in 0.31 seconds on a single NVIDIA A6000 GPU, demonstrating low-latency generation under its causal formulation. 
Further acceleration via KV caching is feasible and left for future work.

%% file: tabs/main-results.tex
\begin{table*}[t]
\setlength{\tabcolsep}{0.5pt}
\caption{
\textbf{Comparison with DualTalk~\cite{peng2025dualtalk} under progressive-context streaming inference.}  
Each turn corresponds to a $1$-second segment, and the agent’s 3D head motion is predicted 
using $n$ previous turns as \textit{context history} ($n=0,3,7$), where $n=0$ denotes no history. 
Metrics with $\downarrow$ indicate lower is better (FD, P-FD, MSE, rPCC), 
and $\uparrow$ indicates higher is better (SID). 
DualTalk$^{*}$ denotes the official checkpoint, and DualTalk$^{\dagger}$ our re-trained model. 
\textcolor{green!10}{\rule{1.5em}{0.8em}} indicates improvement over the best-performing metric, 
while \textcolor{red!10}{\rule{1.5em}{0.8em}} denotes a drop or no change.
}
\label{tab:main-results}
\resizebox{\linewidth}{!}{
\begin{tabular}{
l
>{\columncolor{gray!12}}c
>{\columncolor{gray!12}}c
>{\columncolor{gray!12}}c
c
c
c
>{\columncolor{gray!12}}c
>{\columncolor{gray!12}}c
>{\columncolor{gray!12}}c
c
c
c
>{\columncolor{gray!12}}c
>{\columncolor{gray!12}}c
>{\columncolor{gray!12}}c
}
& \multicolumn{3}{c}{\textbf{FD} $\downarrow$} & \multicolumn{3}{c}{\textbf{P-FD} $\downarrow$}   & \multicolumn{3}{c}{\textbf{MSE} $\downarrow$} & \multicolumn{3}{c}{\textbf{SID} $\uparrow$}  & \multicolumn{3}{c}{\textbf{rPCC} $\downarrow$} \\
\rowcolor{gray!40}
\headcell{Methods}{} &
\headcell{EXP}{} &
\headcell{JAW}{$\times10^3$} &
\headcell{POSE}{$\times10^2$} &
\headcell{EXP}{} &
\headcell{JAW}{$\times10^3$} &
\headcell{POSE}{$\times10^2$} &
\headcell{EXP}{$\times10^1$} &
\headcell{JAW}{$\times10^3$} &
\headcell{POSE}{$\times10^2$} &
\headcell{EXP}{} &
\headcell{JAW}{} &
\headcell{POSE}{} &
\headcell{EXP}{$\times10^2$} &
\headcell{JAW}{$\times10^1$} &
\headcell{POSE}{$\times10^1$} \\
 \specialrule{0em}{1pt}{1pt}
 \specialrule{0em}{1pt}{1pt}
\multicolumn{16}{c}{\textbf{\textit{Test Dataset}}} \\
\rowcolor{gray!25}\multicolumn{16}{l}{\textbf{\textit{Context  History  ($n=0$)}}} \\
DualTalk\textsuperscript{*} & 13.93 & 1.90   & 3.42  & 15.33 & 2.02  & 3.70   & 5.35  & 1.53  & 2.27  & 2.99  & 2.14  & 1.69 & 5.34  & 1.33  & 2.28 \\
DualTalk\textsuperscript{$\dagger$} & 14.16 & 1.98  & 3.63  & 15.59 & 2.10  & 3.91  & 5.34  & 1.49  & 2.36  & 2.95  & 2.14  & 1.66  & 5.97  & 1.34  & 2.28 \\
TIMAR & 
9.61\downdiff{-4.32} &
1.51\downdiff{-0.39} &
2.98\downdiff{-0.44} &
10.91\downdiff{-4.42} &
1.63\downdiff{-0.39} &
3.27\downdiff{-0.43} &
3.94\downdiff{-1.40} &
1.15\downdiff{-0.34} &
1.73\downdiff{-0.54} &
3.42\updiff{+0.43} &
2.33\updiff{+0.19} &
1.84\updiff{+0.15} &
4.51\downdiff{-0.83} &
1.19\downdiff{-0.14} &
2.26\downdiff{-0.02} \\
\specialrule{0em}{1pt}{1pt}
\rowcolor{gray!25}\multicolumn{16}{l}{\textbf{\textit{Context  History  ($n=3$)}}} \\
DualTalk\textsuperscript{*} & 11.53 & 1.67  & 3.22  & 12.75 & 1.79  & 3.48  & 4.42  & 1.28  & 2.00  & 3.21  & 2.24  & 1.76  & 4.88  & 1.23  & 2.16 \\
DualTalk\textsuperscript{$\dagger$} & 11.25 & 1.68  & 3.38  & 12.51 & 1.79  & 3.65  & 4.32  & 1.24  & 2.08  & 3.25  & 2.26  & 1.73  & 4.98  & 1.24  & 2.18 \\
TIMAR & 
9.11\downdiff{-2.14} &
1.57\downdiff{-0.10} &
3.06\downdiff{-0.16} &
10.13\downdiff{-2.38} &
1.66\downdiff{-0.13} &
3.28\downdiff{-0.20} &
3.61\downdiff{-0.71} &
1.09\downdiff{-0.15} &
1.63\downdiff{-0.37} &
3.51\updiff{+0.26} &
2.34\updiff{+0.08} &
1.86\updiff{+0.10} &
4.19\downdiff{-0.69} &
1.19\downdiff{-0.04} &
2.19\downdiff{+0.03} \\
\specialrule{0em}{1pt}{1pt}
\rowcolor{gray!25}\multicolumn{16}{l}{\textbf{\textit{Context  History  ($n=7$)}}} \\
DualTalk\textsuperscript{*} & 11.26 & 1.67  & 3.29  & 12.41 & 1.78  & 3.53  & 4.20   & 1.20   & 1.92  & 3.28  & 2.27  & 1.77   & 4.84  & 1.23  & 2.18 \\
DualTalk\textsuperscript{$\dagger$} & 10.92 & 1.69  & 3.43  & 12.09 & 1.79  & 3.68  & 4.09  & 1.18  & 1.99  & 3.31  & 2.26  & 1.74  & 4.81  & 1.25  & 2.20 \\
TIMAR & 
8.97\downdiff{-1.95} &
1.57\downdiff{-0.10} &
3.08\downdiff{-0.21} &
9.93\downdiff{-2.16} &
1.65\downdiff{-0.13} &
3.28\downdiff{-0.25} &
3.58\downdiff{-0.51} &
1.07\downdiff{-0.11} &
1.61\downdiff{-0.31} &
3.53\updiff{+0.22} &
2.35\updiff{+0.08} &
1.85\updiff{+0.08} &
4.12\downdiff{-0.69} &
1.22\downdiff{-0.01} &
2.18\downdiff{+0.00} \\
\specialrule{0em}{1pt}{1pt}
\specialrule{0em}{1pt}{1pt}
\multicolumn{16}{c}{\textbf{\textit{Out-of-Distribution Dataset}}} \\
\rowcolor{gray!25}\multicolumn{16}{l}{\textbf{\textit{Context  History  ($n=0$)}}} \\
DualTalk\textsuperscript{*} & 22.44 & 2.80  & 5.03  & 23.89 & 2.91  & 5.34  & 7.30  & 1.86  & 2.89  & 2.67  & 1.94  & 1.41  & 6.86  & 1.67  & 3.00 \\
DualTalk\textsuperscript{$\dagger$} & 22.73 & 2.71  & 4.82  & 24.22 & 2.81  & 5.12  & 7.28  & 1.77  & 2.83  & 2.64  & 1.98  & 1.45  & 7.11  & 1.55  & 2.98 \\
TIMAR & 
20.62\downdiff{-1.82} &
2.50\downdiff{-0.21} &
4.31\downdiff{-0.51} &
22.10\downdiff{-1.79} &
2.62\downdiff{-0.19} &
4.62\downdiff{-0.50} &
6.46\downdiff{-0.82} &
1.56\downdiff{-0.21} &
2.29\downdiff{-0.54} &
2.85\updiff{+0.18} &
2.03\updiff{+0.05} &
1.55\updiff{+0.10} &
6.76\downdiff{-0.10} &
1.55\downdiff{+0.00} &
2.77\downdiff{-0.21} \\
\specialrule{0em}{1pt}{1pt}
\rowcolor{gray!25}\multicolumn{16}{l}{\textbf{\textit{Context  History  ($n=3$)}}} \\
DualTalk\textsuperscript{*} & 21.33 & 2.72  & 4.96  & 22.64 & 2.82  & 5.25  & 6.70  & 1.69  & 2.73  & 2.76  & 1.97  & 1.44  & 6.78  & 1.66  & 2.94 \\
DualTalk\textsuperscript{$\dagger$} & 21.25 & 2.64  & 4.84  & 22.60 & 2.75  & 5.14  & 6.60  & 1.62  & 2.66  & 2.78  & 2.00  & 1.48  & 6.67  & 1.56  & 2.85 \\
TIMAR & 
20.21\downdiff{-1.04} &
2.49\downdiff{-0.15} &
4.36\downdiff{-0.48} &
21.38\downdiff{-1.22} &
2.60\downdiff{-0.15} &
4.60\downdiff{-0.54} &
6.16\downdiff{-0.44} &
1.49\downdiff{-0.13} &
2.19\downdiff{-0.47} &
2.90\updiff{+0.12} &
2.03\updiff{+0.03} &
1.54\updiff{+0.06} &
6.47\downdiff{-0.20} &
1.56\downdiff{+0.00} &
2.70\downdiff{-0.15} \\
\specialrule{0em}{1pt}{1pt}
\rowcolor{gray!25}\multicolumn{16}{l}{\textbf{\textit{Context  History  ($n=7$)}}} \\
DualTalk\textsuperscript{*} & 21.31 & 2.75  & 5.13  & 22.56 & 2.85  & 5.40  & 6.53  & 1.63  & 2.67  & 2.81  & 1.97  & 1.43  & 6.79  & 1.70  & 3.02 \\
DualTalk\textsuperscript{$\dagger$} & 21.21 & 2.72  & 4.96  & 22.48 & 2.82  & 5.23  & 6.42  & 1.59  & 2.58  & 2.81  & 2.01  & 1.45  & 6.62  & 1.59  & 2.79 \\
TIMAR &
20.23\downdiff{-0.98} &
2.56\downdiff{-0.16} &
4.50\downdiff{-0.46} &
21.34\downdiff{-1.14} &
2.66\downdiff{-0.16} &
4.72\downdiff{-0.51} &
6.17\downdiff{-0.25} &
1.48\downdiff{-0.11} &
2.18\downdiff{-0.40} &
2.93\updiff{+0.12} &
2.04\updiff{+0.03} &
1.52\updiff{+0.07} &
6.26\downdiff{-0.36} &
1.57\downdiff{-0.02} &
2.77\downdiff{-0.02} \\
\end{tabular}
}
\end{table*}

%% file: tabs/other-baselines.tex
\begin{table*}[t]
\caption{
\textbf{Comparison with existing baselines under the standard (non-streaming) DualTalk benchmark protocol.} 
Except for the DualTalk model, all results are taken from the DualTalk paper. 
\textbf{Bold} indicates the best performance, and \underline{underlined} values denote the second best.
Other notations follow the caption of \Cref{tab:main-results}.
}
\label{tab:other-baselines}
\resizebox{\linewidth}{!}{
\begin{tabular}{
l
>{\columncolor{gray!12}}c
>{\columncolor{gray!12}}c
>{\columncolor{gray!12}}c
c
c
c
>{\columncolor{gray!12}}c
>{\columncolor{gray!12}}c
>{\columncolor{gray!12}}c
c
c
c
>{\columncolor{gray!12}}c
>{\columncolor{gray!12}}c
>{\columncolor{gray!12}}c
}
 & 
\multicolumn{3}{c}{\textbf{FD} $\downarrow$} & 
\multicolumn{3}{c}{\textbf{P-FD} $\downarrow$} & 
\multicolumn{3}{c}{\textbf{MSE} $\downarrow$} & 
\multicolumn{3}{c}{\textbf{SID} $\uparrow$} & 
\multicolumn{3}{c}{\textbf{rPCC} $\downarrow$} \\
\rowcolor{gray!40}
\headcell{Methods}{} &
\headcell{EXP}{} &
\headcell{JAW}{$\times10^3$} &
\headcell{POSE}{$\times10^2$} &
\headcell{EXP}{} &
\headcell{JAW}{$\times10^3$} &
\headcell{POSE}{$\times10^2$} &
\headcell{EXP}{$\times10^1$} &
\headcell{JAW}{$\times10^3$} &
\headcell{POSE}{$\times10^2$} &
\headcell{EXP}{} &
\headcell{JAW}{} &
\headcell{POSE}{} &
\headcell{EXP}{$\times10^2$} &
\headcell{JAW}{$\times10^1$} &
\headcell{POSE}{$\times10^1$} \\
 \specialrule{0em}{1pt}{1pt}
 \specialrule{0em}{1pt}{1pt}
\multicolumn{16}{c}{\textbf{\textit{Test Dataset}}} \\
FaceFormer~\cite{fan2022faceformer} & 
34.90 & 
5.40 & 
8.00 & 
34.90 & 
5.40 & 
8.00 & 
6.97 & 
1.80 & 
2.67 & 
0.54 & 
0.36 & 
0.50 & 
13.05 & 
2.41 & 
5.27 \\
CodeTalker~\cite{xing2023codetalker} & 
48.57 & 
6.89 & 
10.74 & 
48.57 & 
6.89 & 
10.74 & 
9.71 & 
2.29 & 
3.58 & 
0 & 
0 & 
0 & 
11.06 & 
2.33 & 
5.11 \\
EmoTalk~\cite{peng2023emotalk} & 
29.86 & 
4.33 & 
7.54 & 
30.20 & 
4.36 & 
7.58 & 
6.88 & 
1.76 & 
2.59 & 
2.86 & 
1.72 & 
0.98 & 
9.89 & 
2.19 & 
4.94 \\
SelfTalk~\cite{peng2023selftalk} & 
35.77 & 
5.49 & 
8.14 & 
35.77 & 
5.49 & 
8.14 & 
7.15 & 
1.83 & 
2.71 & 
2.49 & 
1.30 & 
1.28 & 
12.25 & 
2.39 & 
4.70 \\
L2L~\cite{ng2022learning} & 
24.61 & 
3.69 & 
7.08 & 
24.99 & 
3.74 & 
7.13 & 
5.68 & 
1.48 & 
2.49 & 
2.86 & 
1.89 & 
1.19 & 
8.52 & 
2.06 & 
4.11  \\
DualTalk\textsuperscript{*}~\cite{peng2025dualtalk}  & 
11.14 & 
\underline{1.90} & 
\underline{3.83} & 
11.88 & 
\underline{1.97} & 
\underline{3.97} & 
\underline{3.59} & 
\textbf{1.04} & 
\underline{1.72} & 
3.48 & 
2.23 & 
\underline{1.72} & 
4.73 & 
1.37 & 
\underline{2.38} \\
DualTalk\textsuperscript{$\dagger$}~\cite{peng2025dualtalk} & 
\underline{11.08} & 
1.97  & 
4.03  & 
\underline{11.82} & 
2.03  & 
4.17  & 
\textbf{3.52}  & 
\textbf{1.04}  & 
1.78  & 
\underline{3.50}  & 
\underline{2.25}  & 
1.70  & 
\underline{4.62}  & 
\underline{1.35}  & 
2.45 \\
TIMAR &
\textbf{8.91}  & 
\textbf{1.57}  & 
\textbf{3.06}  & 
\textbf{9.88}  & 
\textbf{1.65}  & 
\textbf{3.26}  & 
3.60  & 
\underline{1.07}  & 
\textbf{1.61} & 
\textbf{3.55}  & 
\textbf{2.36}  & 
\textbf{1.87}  & 
\textbf{4.10}  & 
\textbf{1.22}  & 
\textbf{2.17} \\
\specialrule{0pt}{2pt}{2pt}
\multicolumn{16}{c}{\textbf{\textit{Out-of-Distribution Dataset}}} \\
FaceFormer~\cite{fan2022faceformer} & 
35.92 & 
5.39 & 
8.60 & 
35.93 & 
5.39 & 
8.60 & 
7.18 & 
1.80 & 
2.87 & 
0.54 & 
0.40 & 
0.51 & 
11.71 & 
2.16 & 
5.73  \\
CodeTalker~\cite{xing2023codetalker} & 
50.05 & 
6.95 & 
11.66 & 
50.05 & 
6.95 & 
11.66 & 
10.01 & 
2.32 & 
3.88 & 
0 & 
0 & 
0 & 
10.24 & 
2.18 & 
5.76  \\
EmoTalk~\cite{peng2023emotalk} & 
34.12 & 
4.17 & 
8.59 & 
34.44 & 
4.21 & 
8.62 & 
7.73 & 
1.71 & 
2.94 & 
2.89 & 
1.79 & 
0.94 & 
9.44 & 
1.96 & 
5.54                                                 \\
SelfTalk~\cite{peng2023selftalk} & 
36.23 & 
5.36 & 
8.89 & 
36.23 & 
5.36 & 
8.89 & 
7.24 & 
1.79 & 
2.96 & 
2.61 & 
1.36 & 
1.08 & 
11.26 & 
2.13 & 
5.67                                                 \\
L2L~\cite{ng2022learning} & 
30.49 & 
3.82 & 
8.56 & 
30.87 & 
3.86 & 
8.61 & 
6.87 & 
1.54 & 
2.98 & 
2.76 & 
1.91 & 
1.11 & 
9.02 & 
1.94 & 
4.99                                                 \\
DualTalk\textsuperscript{*}~\cite{peng2025dualtalk} & 
\underline{21.71} & 
\underline{3.15} & 
\underline{5.89} & 
\underline{22.56} & 
\underline{3.22} & 
\underline{6.06} & 
\underline{5.97} & 
\underline{1.50} & 
2.48 & 
\underline{2.98} & 
1.94 & 
\underline{1.38} & 
6.86 & 
\underline{1.60} & 
3.28                                        \\
DualTalk\textsuperscript{$\dagger$}~\cite{peng2025dualtalk} & 
21.91 & 
3.20  & 
5.94  & 
22.71 & 
3.27  & 
6.11  & 
\textbf{5.89}  & 
\underline{1.50}  & 
\underline{2.46}  & 
\textbf{3.01}  & 
\underline{1.96}  & 
1.33 & 
\underline{6.48}  & 
1.77  & 
\underline{3.05} \\
TIMAR &
\textbf{20.20} & 
\textbf{2.56}  & 
\textbf{4.48}  & 
\textbf{21.33} & 
\textbf{2.66}  & 
\textbf{4.70}  & 
6.20  & 
\textbf{1.48}  & 
\textbf{2.18}  & 
2.96  & 
\textbf{2.04}  & 
\textbf{1.53}  & 
\textbf{6.22}  & 
\textbf{1.55}  & 
\textbf{2.76} \\
\end{tabular}
}
\end{table*}

%% file: figs/qulitative_comp_context.tex
\begin{figure*}[t]
\centering
\includegraphics[width=1\linewidth]{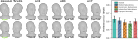}
\caption{
\textbf{Case study and user study comparison with DualTalk.}
\textit{\textbf{Left:}} Progressive-context streaming results ($n{=}0,3,7$); 
green \textit{agent} denotes the ground-truth reference and \textit{user} the interlocutor. 
\textit{\textbf{Right:}} User preference rates for TIMAR over DualTalk (ties counted as 0.5) with 95\% bootstrap confidence intervals; 
the dashed line indicates no preference (0.5).
Implementation details are provided in Appendix~\cref{app:user-study}.
}
\label{fig:qulitative_comp_context}
\end{figure*}

%% file: figs/cfg_case.tex
\begin{figure*}[t]
\centering
\includegraphics[width=1\linewidth]{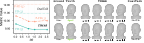}
\caption{
\textbf{Effect of classifier-free guidance (CFG) during sampling.}
\textit{\textbf{Left:}} Quantitative results showing FD and P-FD metrics under different guidance scales $\omega$.
\textit{\textbf{Right:}} Visual comparison of generated agent heads with varying $\omega$, where higher guidance improves contextual consistency and expressiveness.
DualTalk cannot support CFG-based sampling.
Green text denotes the agent, whose 3D head is predicted.
}
\label{fig:cfg}
\end{figure*}

%% file: figs/param_perf--ablation-diff-mlp-loss.tex
\begin{figure}[t]
\centering

\begin{minipage}{0.49\linewidth}
\centering
\includegraphics[width=\linewidth]{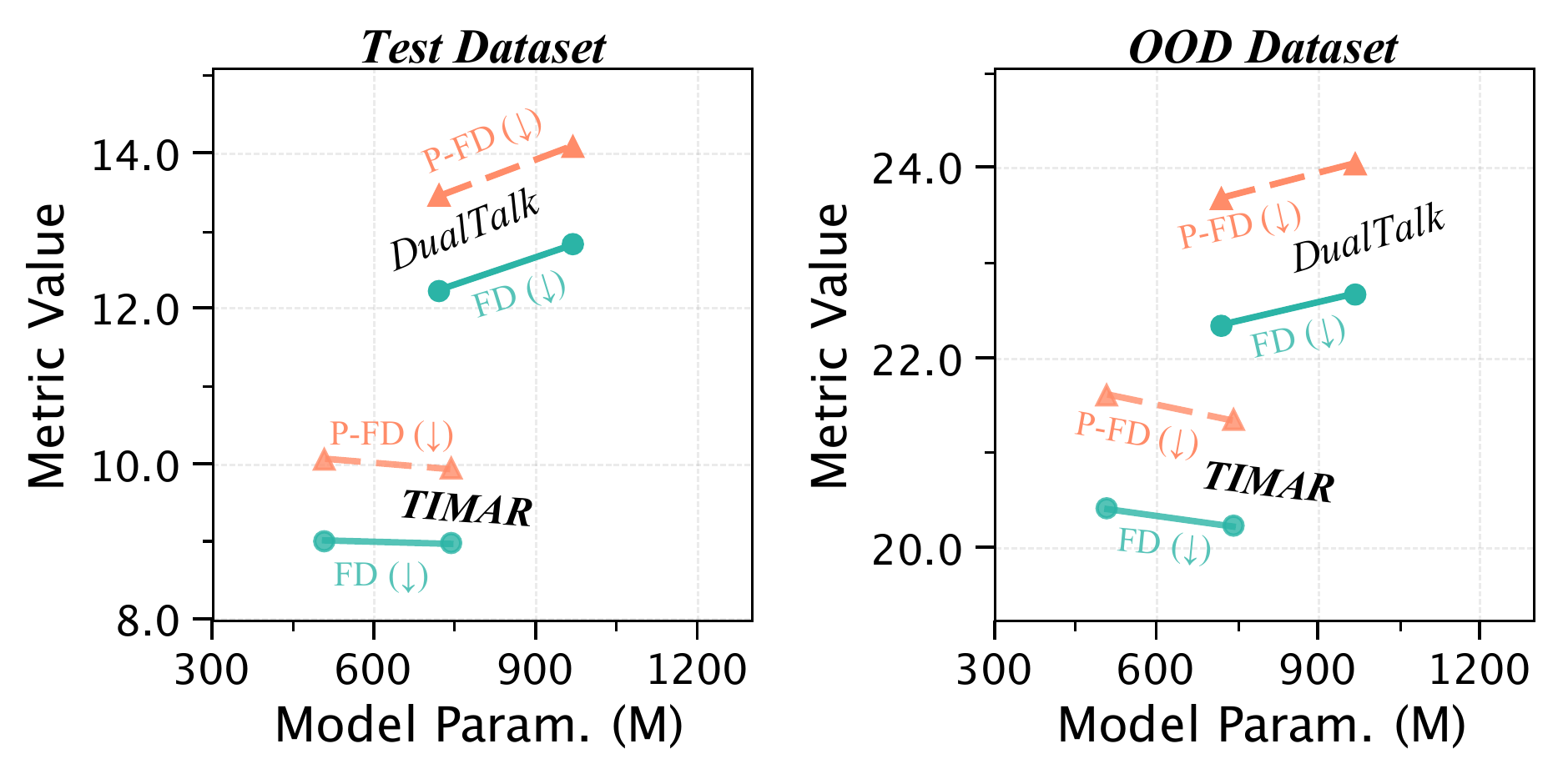}
\caption{
\textbf{Model performance versus parameter size.}  
FD and P-FD metrics (\textit{lower is better}) are computed on the \textit{test} dataset for the 3D head \textit{EXP} parameters.  
Results show that enlarging the DualTalk model does not improve performance, whereas TIMAR achieves lower errors with fewer or comparable parameters.
}
\label{fig:param-perf}
\end{minipage}
\hfill
\begin{minipage}{0.49\linewidth}
\centering
\includegraphics[width=\linewidth]{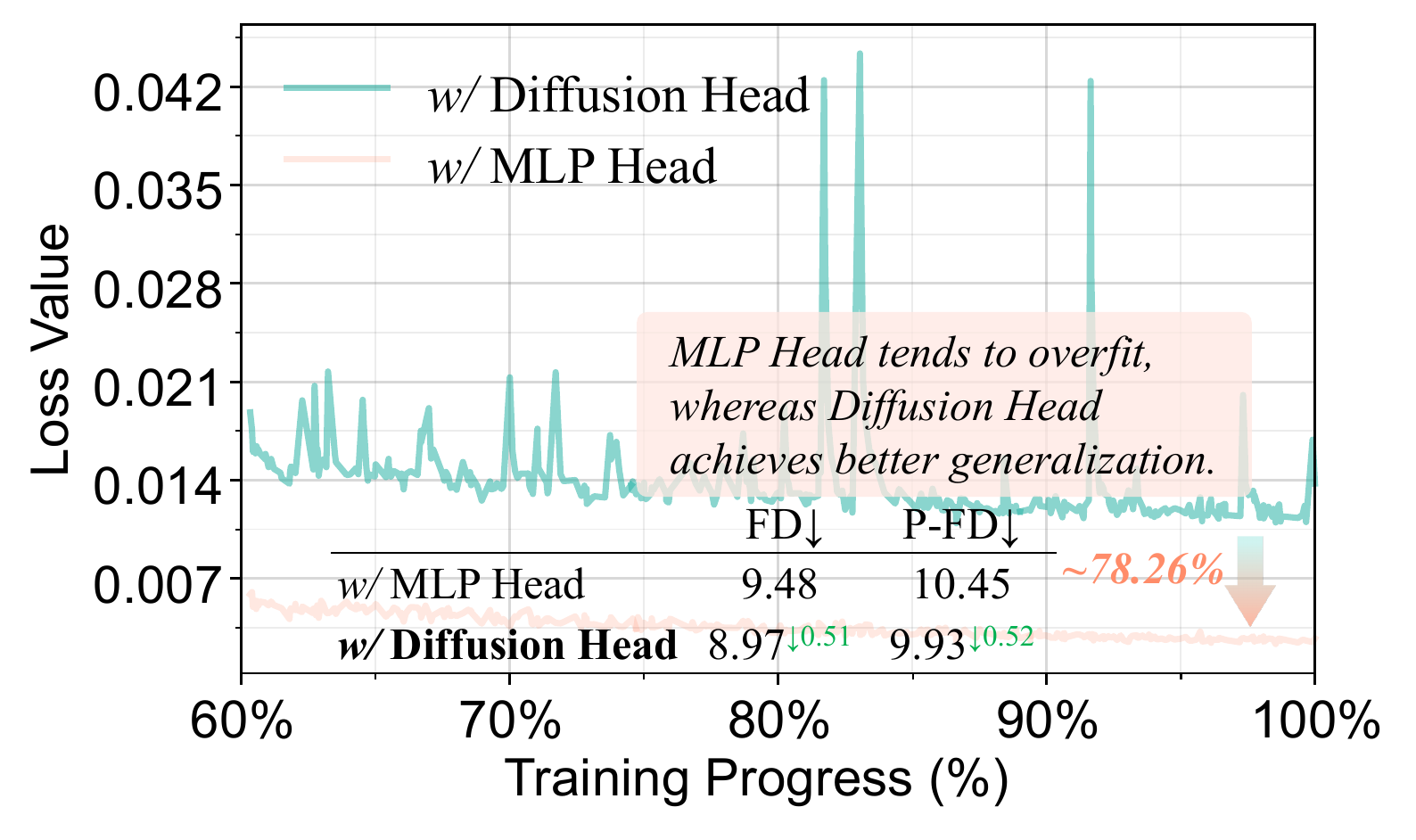}
\caption{
\textbf{Comparison of training dynamics between Diffusion Head and MLP Head.}  
Both variants are trained under identical settings, and their loss curves are directly comparable.  
Detailed quantitative results are provided in \Cref{tab:main-ablation}.
}
\label{fig:ablation-diffusion-head-mlp-head-loss}
\end{minipage}

\end{figure}

%% file: tabs/main-ablation.tex
\begin{table*}[t]
\setlength{\tabcolsep}{1.5pt}
\caption{
\textbf{Ablation study on core architectural components.}  
We compare TIMAR with alternative designs across three aspects:  
(i) replacing the diffusion-based head with a direct MLP predictor (\textit{Bottleneck Ablation}),  
(ii) substituting the proposed Turn-Level Causal Attention (TLCA) with full bidirectional attention (\textit{Attention Ablation}),  
and (iii) adopting an asymmetric encoder–decoder design following MAE~\cite{he2021masked} instead of the encoder-only backbone (\textit{Backbone Architecture Ablation}).  
Results are reported on the \emph{test} dataset.
}
\label{tab:main-ablation}
\resizebox{\linewidth}{!}{
\begin{tabular}{
l
>{\columncolor{gray!12}}c
>{\columncolor{gray!12}}c
>{\columncolor{gray!12}}c
c
c
c
>{\columncolor{gray!12}}c
>{\columncolor{gray!12}}c
>{\columncolor{gray!12}}c
c
c
c
>{\columncolor{gray!12}}c
>{\columncolor{gray!12}}c
>{\columncolor{gray!12}}c
}
& \multicolumn{3}{c}{\textbf{FD} $\downarrow$} & \multicolumn{3}{c}{\textbf{P-FD} $\downarrow$}   & \multicolumn{3}{c}{\textbf{MSE} $\downarrow$} & \multicolumn{3}{c}{\textbf{SID} $\uparrow$}  & \multicolumn{3}{c}{\textbf{rPCC} $\downarrow$} \\
\rowcolor{gray!40}
\headcell{Methods}{} &
\headcell{EXP}{} &
\headcell{JAW}{$\times10^3$} &
\headcell{POSE}{$\times10^2$} &
\headcell{EXP}{} &
\headcell{JAW}{$\times10^3$} &
\headcell{POSE}{$\times10^2$} &
\headcell{EXP}{$\times10^1$} &
\headcell{JAW}{$\times10^3$} &
\headcell{POSE}{$\times10^2$} &
\headcell{EXP}{} &
\headcell{JAW}{} &
\headcell{POSE}{} &
\headcell{EXP}{$\times10^2$} &
\headcell{JAW}{$\times10^1$} &
\headcell{POSE}{$\times10^1$} \\
\specialrule{0em}{4pt}{4pt}
\rowcolor{gray!25}\multicolumn{16}{l}{\textbf{\textit{Bottleneck Ablation}}} \\
\textit{w/} MLP Head & 9.48  & 1.67  & 3.36  & 10.45 & 1.76  & 3.56  & 3.43  & 1.12  & 1.71  & 3.50   & 2.34  & 1.77  & 4.40   & 1.38  & 2.41 \\
\textbf{\textit{w/} Diffusion Head} & 
8.97\downdiff{-0.51} & 
1.57\downdiff{-0.10} & 
3.08\downdiff{-0.28} & 
9.93\downdiff{-0.52} & 
1.65\downdiff{-0.11} & 
3.28\downdiff{-0.28} & 
3.58\downdiff{0.15} & 
1.07\downdiff{-0.05} & 
1.61\downdiff{-0.10} & 
3.53\updiff{0.03} & 
2.35\updiff{0.01} & 
1.85\updiff{0.08} & 
4.12\downdiff{-0.28} & 
1.22\downdiff{-0.16} & 
2.18\downdiff{-0.23} \\
\specialrule{0em}{1pt}{1pt}
\rowcolor{gray!25}\multicolumn{16}{l}{\textbf{\textit{Attention Ablation}}} \\
\textit{w/} Bi-Attention & 9.12  & 1.82  & 3.04  & 10.13 & 1.91  & 3.25  & 3.67  & 1.18  & 1.62  & 3.54  & 2.27  & 1.86  & 4.12  & 1.40   & 2.21 \\
\textbf{\textit{w/} TLCA} & 
8.97\downdiff{-0.15} & 
1.57\downdiff{-0.25} & 
3.08\downdiff{0.04} & 
9.93\downdiff{-0.20} & 
1.65\downdiff{-0.26} & 
3.28\downdiff{0.03} & 
3.58\downdiff{-0.09} & 
1.07\downdiff{-0.11} & 
1.61\downdiff{-0.01} & 
3.53\updiff{-0.01} & 
2.35\updiff{0.08} & 
1.85\updiff{-0.01} & 
4.12\downdiff{0.00} & 
1.22\downdiff{-0.18} & 
2.18\downdiff{-0.03} \\
\specialrule{0em}{1pt}{1pt}
\rowcolor{gray!25}\multicolumn{16}{l}{\textbf{\textit{Backbone Architecture Ablation}}} \\
\textit{w/} Asym. EncDec & 9.51  & 1.75  & 3.19  & 10.43 & 1.84  & 3.38  & 3.57  & 1.13  & 1.63  & 3.51  & 2.30   & 1.85  & 4.15  & 1.31  & 2.19 \\
\textbf{\textit{w/} Encoder-Only} & 
8.97\downdiff{-0.54} & 
1.57\downdiff{-0.18} & 
3.08\downdiff{-0.11} & 
9.93\downdiff{-0.50} & 
1.65\downdiff{-0.19} & 
3.28\downdiff{-0.10} & 
3.58\downdiff{0.01} & 
1.07\downdiff{-0.06} & 
1.61\downdiff{-0.02} & 
3.53\updiff{0.02} & 
2.35\updiff{0.05} & 
1.85\updiff{0.00} & 
4.12\downdiff{-0.03} & 
1.22\downdiff{-0.09} & 
2.18\downdiff{-0.01} \\
\end{tabular}
}
\end{table*}

%% file: figs/robustness.tex
\begin{figure}[t]
\centering
\begin{minipage}[t]{0.48\linewidth}
\centering
\includegraphics[width=\linewidth]{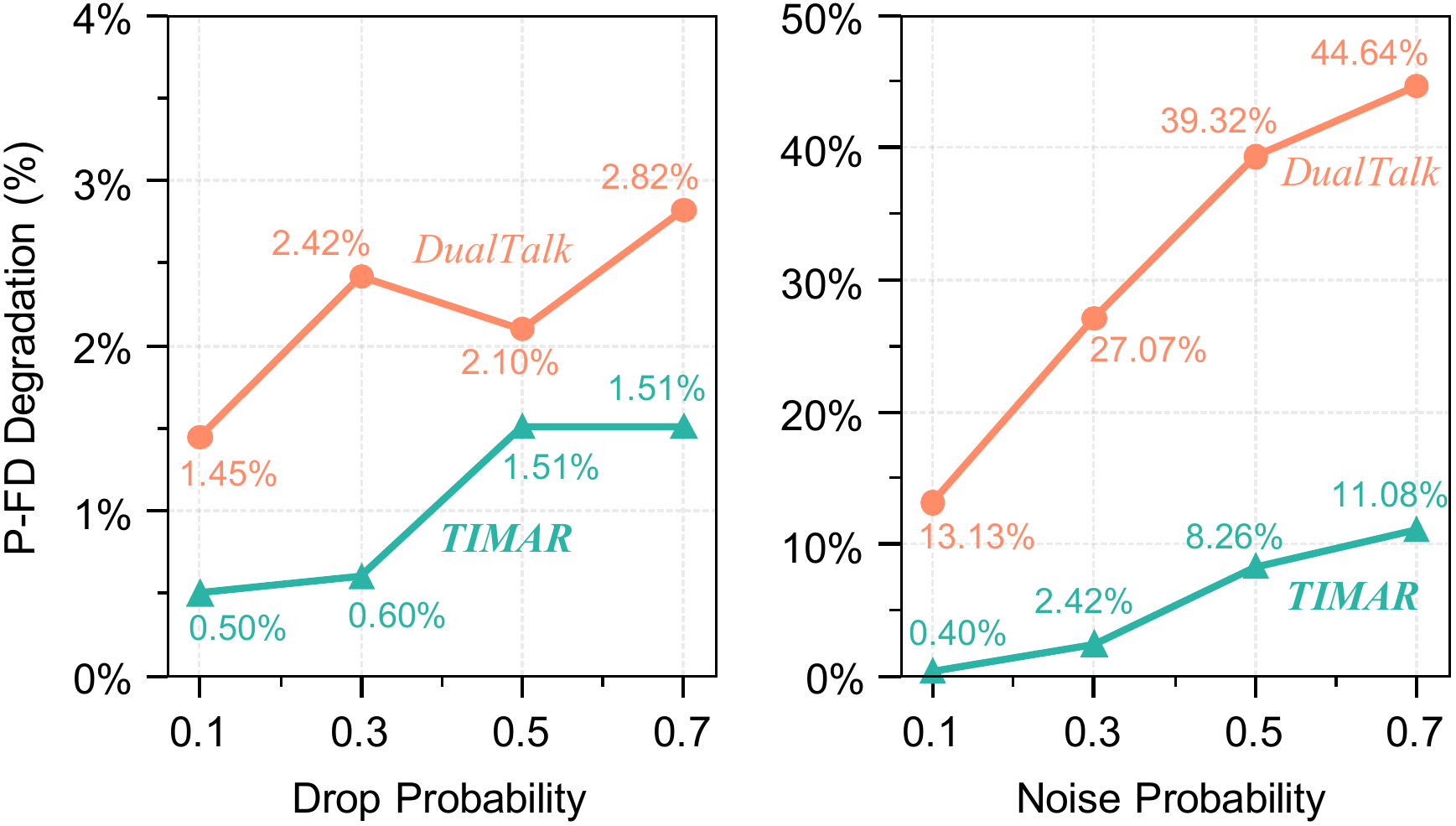}
\caption{Robustness under head corruption}
\label{fig:head_motion_failure}
\end{minipage}\hfill
\begin{minipage}[t]{0.48\linewidth}
\centering
\includegraphics[width=\linewidth]{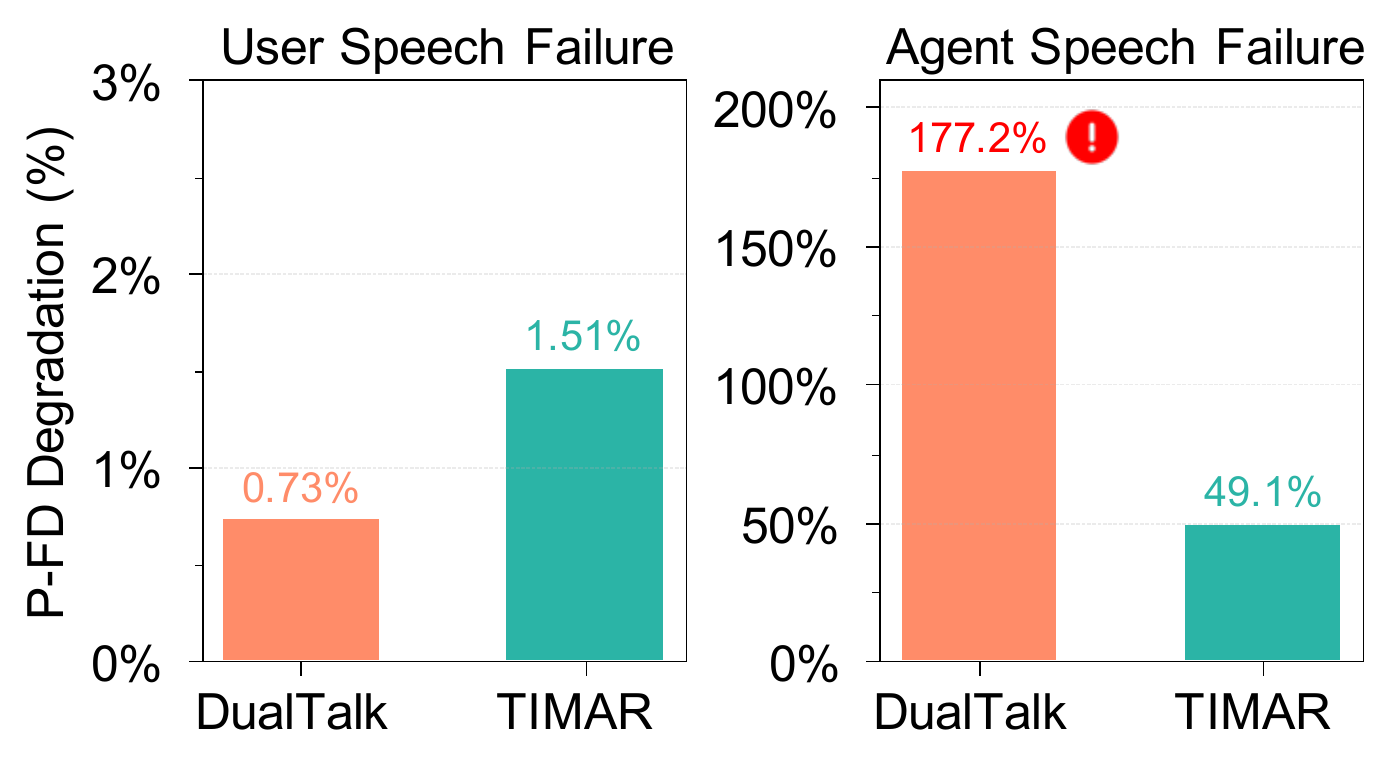}
\caption{Robustness under speech failures}
\label{fig:speech_failure}
\end{minipage}
\end{figure}

%% file: sec/5_conclusion.tex
\section{Conclusion}
\label{sec:conclusion}
We introduced \textbf{TIMAR}, a causal framework for interactive 3D conversational head generation with turn-level autoregressive diffusion.  
By interleaving multimodal tokens under causal attention, TIMAR models cross-speaker dependencies and preserves temporal coherence.  
Experiments show consistent improvements over DualTalk in realism and synchronization, and ablations validate the diffusion head and causal design.  
Overall, explicit causal and interleaved modeling leads to more humanlike conversational dynamics.

%% file: sec/X_suppl.tex
\clearpage
\appendix
\setcounter{page}{1}
\title{Supplementary Material for ``TIMAR: Causal Turn-Level Modeling of Interactive 3D Conversational Head Dynamics''}
\maketitle


\section{Problem Statement}\label{app:problem-statement}

We study the task of \textbf{interactive 3D conversational head generation} in dyadic settings. 
Given two participants, a \textit{user} and an \textit{agent}, the goal is to synthesize the agent's 3D head motion that coherently reflects both speaking and listening behaviors. 
The generated motion should capture verbal articulation (\eg, lip and jaw movements) as well as nonverbal feedback (\eg, nodding, gaze shifts, and subtle expressions), conditioned on the evolving multimodal conversational context.
We represent head motion in a parametric 3DMM space, consistent with prior interactive conversational modeling work such as DualTalk~\cite{peng2025dualtalk}, which provides a compact and semantically structured representation of expression, jaw, and pose.
Unlike purely implicit rendering representations, 3DMM parameters are interpretable, temporally stable, and directly controllable. 
This structured space is particularly suitable for causal modeling of interaction dynamics, and can further serve as a motion layer driving photorealistic neural avatars (\eg, Gaussian-based heads)~\cite{he2025lam} or embodied robotic platforms~\cite{hu2026learning,ze2025twist}. 
Thus, our formulation complements neural rendering approaches by focusing on upstream conversational motion modeling rather than downstream appearance synthesis.

\subsubsection{Signals and Objective.}

Let the user's speech and head motion be $S^{u}$ and $H^{u}$, and the agent's speech be $S^{a}$. 
The target is the agent's head motion $H^{a}$. 
Speech sequences are sampled at rate $f_{s}$ and head motion at frame rate $f_{h}$, 
with $S^{u}, S^{a}\!\in\!\mathbb{R}^{T f_{s}}$ and $H^{u}, H^{a}\!\in\!\mathbb{R}^{T f_{h}\times d_{h}}$, 
where $T$ denotes duration and $d_{h}$ the dimensionality of the 3D head parameters.  
The goal is to model the conditional distribution
\begin{equation}
    p_{\theta}\!\left(H^{a} \mid S^{u}, H^{u}, S^{a}\right),
\end{equation}
which captures plausible 3D head dynamics of the agent given the multimodal context of both speakers.

\subsubsection{Turn-Level Causal Formulation.}

Human conversations evolve sequentially across turns, where behaviors depend on accumulated interaction history rather than future information~\cite{skantze2021turn}. 
However, existing frameworks such as DualTalk~\cite{peng2025dualtalk} rely on bidirectional encoders and full-sequence attention, exposing future frames during training and breaking causal consistency. 
Such formulations are inherently offline and unsuitable for streaming or autoregressive response.

We therefore reformulate the task as a \textbf{turn-level causal generation problem}. 
A dialogue is divided into $N$ fixed-length turns. 
For each turn $t \in \{1,\dots,N\}$, the model observes the user's speech and motion $(S^{u}_{1:t}, H^{u}_{1:t})$ and the agent's speech $(S^{a}_{1:t})$, 
and predicts the agent's head motion at the current turn. 
The joint distribution factorizes causally as
\begin{equation}
\begin{split}
p_{\theta}\!\left(H^{a}_{1:N} \mid S^{u}_{1:N}, H^{u}_{1:N}, S^{a}_{1:N}\right)
= \prod_{t=1}^{N} 
p_{\theta}\!\left(H^{a}_{t} \mid S^{u}_{1:t}, H^{u}_{1:t}, S^{a}_{1:t}\right).
\end{split}
\end{equation}

This causal factorization prevents future leakage and enables turn-by-turn streaming generation, 
while preserving the interleaved structure of dual-speaker interaction so that rhythm, affect, and timing propagate across turns. 
These principles form the foundation of TIMAR.

\section{TIMAR Details}\label{app:timar_details}

\subsection{Network Details}

\subsubsection{Speech Tokenizer.}
We employ the wav2vec~2.0~\cite{baevski2020wav2vec} model as the speech feature extractor $\mathcal{M}_{\text{speech}}$\footnote{We use the \texttt{facebook/wav2vec2-large-960h-lv60-self} checkpoint from the HuggingFace Hub~\cite{Wolf_Transformers_State-of-the-Art_Natural_2020}.}.  
As shown in \Cref{fig:f_speech}, each $c$-second audio chunk $S_{i}\!\in\!\mathbb{R}^{c f_{s}}$ is first passed through the frozen \textit{feature extractor} of wav2vec~2.0,  
which converts raw waveforms into low-frequency acoustic embeddings of dimension $d'_{s}=512$ at a frame rate of $f_{w}=50$~Hz,  
yielding an output of size $\mathbb{R}^{(c f_{w}-1)\times d'_{s}}$.  
These features are linearly interpolated to match the 3D head motion frame rate $f_{h}$, producing $\mathbb{R}^{c f_{h}\times d'_{s}}$ representations.  
The interpolated sequence is then passed through the pretrained \textit{feature projection} and \textit{encoder} modules of $\mathcal{M}_{\text{speech}}$,  
resulting in contextualized embeddings of dimension $d_{s}$.  
Finally, a learnable linear projection $\mathcal{P}_{\text{speech}}:\mathbb{R}^{d_{s}}\!\rightarrow\!\mathbb{R}^{d_{t}}$ maps the features into the shared token space,  
yielding the final token sequence $\mathcal{S}_{i}\!\in\!\mathbb{R}^{c f_{h}\times d_{t}}$ for each chunk.

\subsubsection{3D Head Motion Encoder.}
As shown in \Cref{fig:f_head}, each $c$-second motion segment 
$H_{i}\!\in\!\mathbb{R}^{c f_{h}\times d_{h}}$ consists of $c f_{h}$ frames of 3D head parameters, 
where $d_{h}$ denotes the dimensionality of the FLAME-based~\cite{li2017learning} head representation.  
We implement the motion encoder $\mathcal{F}_{\text{head}}$ as a two-layer multilayer perceptron (MLP) 
with ReLU activations and a hidden dimension of $\tfrac{d_{t}}{2}$,  
followed by a final linear projection to the shared token space of dimension $d_{t}$.  
The encoded feature sequence is denoted as $\mathcal{H}_{i}=\mathcal{F}_{\text{head}}(H_{i})\!\in\!\mathbb{R}^{c f_{h}\times d_{t}}$.

\input{figs/f_speech--f_head}

\subsubsection{Turn-Level Causal Multimodal Fusion.}
As illustrated in \Cref{fig:f_fusion}, the interleaved context sequence 
$\mathcal{T}\!\in\!\mathbb{R}^{N(4c f_{h}+10)\times d_{t}}$ 
consists of $N$ conversational turns,  
where each turn contains four modality-specific token sequences 
(user speech, agent speech, user head, and agent head),  
each representing a $c$-second temporal window at frame rate $f_{h}$.  
The interleaved multimodal sequence is first linearly projected to the Transformer Encoder input dimension $d_{e}$ 
and augmented with a set of learnable \emph{separator tokens}.  
Specifically, ten special tokens are inserted between different modalities 
(\textit{i.e.}, user speech, agent speech, user head, and agent head) 
and between adjacent turns.  
These tokens act as soft delimiters that help the model distinguish modality boundaries 
and prevent temporal leakage across turns,  
while also providing explicit structural cues that stabilize causal attention during training.  
This design choice preserves the turn-level temporal order 
and improves multimodal alignment without altering the causal formulation.

After token augmentation, the sequence is normalized and enriched with a learnable positional embedding $P_{1}$, 
enabling both intra-turn and inter-turn temporal reasoning.  
The Transformer encoder $\mathcal{E}$ equipped with Turn-Level Causal Attention (TLCA) 
then processes the sequence under strict causal masking, 
allowing bidirectional interaction within each turn while constraining cross-turn attention to past tokens only.  
The fused representation $\mathcal{Z}$ encodes both fine-grained multimodal correspondence 
and long-range conversational dependencies, 
serving as the contextual backbone for diffusion-based head generation.

\subsubsection{Lightweight Diffusion Head.}
As shown in \Cref{fig:f_diff}, the diffusion head $\mathcal{F}_{\text{diff}}$ takes the noisy 3D head parameter $x_{t}$ 
and the contextual condition $\mathbf{z}^{m}$ as inputs.  
Both are first linearly projected into a hidden diffusion space of dimension $d_{m}$, 
where $\mathbf{z}^{m}$ is augmented with learnable positional and timestep embeddings to encode frame-level and temporal information.  
The denoising network consists of $K$ residual modulation blocks, 
each performing feature-wise conditional transformation driven by $\mathbf{z}^{m}$.  
For each normalized activation $x$, a modulation operation is applied as:
\begin{equation}
\text{Modulate}(x, \text{shift}, \text{scale}) = x \times (1 + \text{scale}) + \text{shift},
\end{equation}
where \textit{shift} and \textit{scale} are obtained from linear projections of $\mathbf{z}^{m}$.  
This affine modulation allows contextual cues to adaptively rescale and shift the feature responses, 
enabling expressive and stable conditional denoising.  

After passing through $K$ residual modulation blocks, 
the final feature is projected back to the original 3D head parameter space to yield $\hat{\mathbf{h}}^{a}$.  
Despite its compact design, this head effectively captures multimodal stochasticity and preserves temporal coherence in generated 3D motion.

\input{figs/f_fusion--f_diff}

\subsection{Implementation Details}
\subsubsection{Software Framework.}
All experiments are implemented in \texttt{PyTorch} framework.  
Pretrained components, including the wav2vec~2.0 speech tokenizer, are loaded via the \texttt{Transformers} library~\cite{Wolf_Transformers_State-of-the-Art_Natural_2020}.  

\subsubsection{Loss Formulation.}
The 3D head is represented using 56-dimensional FLAME parameters,  
including 50 expression coefficients, 3 jaw, and 3 head pose parameters.  
During training, the diffusion loss $\mathcal{L}_{\text{diff}}$ is computed separately for each subset  
(\textit{i.e.}, expression, jaw, and pose) and then aggregated.  
This separation stabilizes optimization by accounting for the distinct dynamic ranges and semantic sensitivities across different head components.

\subsubsection{Default Configuration.}
The shared token dimension is set to $d_{t}=1024$.
The Transformer encoder in the fusion module contains 16 layers with dimension $d_{e}=1024$ and 16 attention heads.
The diffusion head uses $K=3$ residual modulation blocks, each operating in latent diffusion space $d_{m}=1024$.

\subsubsection{Training Configuration.}
The model is optimized using AdamW~\cite{loshchilov2018decoupled} with a batch size of 32 and 400 total epochs.  
The learning rate is set to $1\times10^{-4}$ with a 100-iteration warm-up schedule.  
Training data are segmented into $T=8$~s clips, where each turn corresponds to a $c=1$~s chunk of temporally aligned user and agent audio-visual streams.  
Speech signals are sampled at $f_{s}=16$~kHz and head motion sequences at $f_{h}=25$~fps.  
During training, 70\% of the agent head tokens are randomly masked ($r=0.7$),  
and classifier-free guidance employs a conditional dropout probability of $p_{\text{cfg}}=0.1$.  

\section{DualTalk Benchmark Details} 

\input{tabs/dataset}

\subsection{Datasets}\label{app:dataset_details}
The DualTalk benchmark dataset provides a large-scale corpus for studying dual-speaker 3D conversational head generation.  
It contains multi-round face-to-face interactions featuring synchronized audio-visual recordings of both participants.  
All videos are sourced from open-domain interview and dialogue recordings, selected to ensure clear frontal visibility of both speakers and high-quality audio tracks.  
Each video is recorded at $1920{\times}1080$ resolution and $25$~fps, with audio sampled at $16$~kHz.  
Speaker separation, tracking, and 3D reconstruction are performed following the official DualTalk preprocessing pipeline to obtain temporally aligned 3D head parameters and speech signals for both participants.
The released dataset comprises approximately $50$ hours of processed conversation data, covering more than $1000$ distinct identities and $5763$ conversational samples in total.  
The official data split includes $4853$ samples for training, $533$ for testing, and $377$ for out-of-distribution (OOD) evaluation.  
The OOD set contains unseen speakers and conversation scenarios to assess generalization.  
Table~\ref{tab:dataset_stats} summarizes the overall data scale and the distribution of conversation rounds, where most dialogues contain one to three alternating speaker turns, reflecting natural short-turn interaction patterns.

\subsection{Metrics}\label{app:metric_details}
\subsubsection{Fréchet Distance (FD).}
FD measures the distributional similarity between generated and ground-truth motions in a deep feature space.  
Given activation statistics $(\mu_{1}, \Sigma_{1})$ and $(\mu_{2}, \Sigma_{2})$ from a pretrained encoder,  
it is computed as
\[
\text{FD} = \|\mu_{1}-\mu_{2}\|^{2} + \operatorname{Tr}\!\left(\Sigma_{1}+\Sigma_{2}-2(\Sigma_{1}\Sigma_{2})^{\frac{1}{2}}\right),
\]
where lower values indicate closer alignment between the generated and real motion distributions.

\subsubsection{Paired Fréchet Distance (P-FD).}
P-FD extends FD to paired motion embeddings by concatenating generated agent motion with corresponding user motion before computing the distance.
This paired variant evaluates how well generated motion maintains inter-speaker coherence and synchronization.

\subsubsection{Mean Squared Error (MSE).}
MSE quantifies frame-level reconstruction accuracy between predicted and ground-truth 3D head parameters:
\[
\text{MSE} = \frac{1}{N}\sum_{i=1}^{N}\| \hat{\mathbf{h}}_{i} - \mathbf{h}_{i} \|^{2}.
\]
Lower MSE values indicate higher fidelity to the reference.

\subsubsection{SI for Diversity (SID).}
SID measures the diversity of generated motion.  
Following DualTalk, $k$-means clustering ($k=40$) is applied to motion features,  
and the entropy of the cluster assignment histogram is computed as
\[
\text{SID} = - \sum_{k=1}^{K} p_{k}\log_{2}(p_{k}+\epsilon),
\]
where $p_{k}$ denotes the normalized cluster occupancy.  
Higher SID indicates greater motion variety and less repetition.

\subsubsection{Residual Pearson Correlation Coefficient (rPCC).}
rPCC evaluates the temporal correlation between user and agent behaviors.  
It computes the Pearson correlation of motion trajectories for each speaker pair and measures the L1 distance between the generated and real correlation patterns.  
Lower rPCC values correspond to more accurate modeling of interactive timing and responsiveness.

\subsubsection{Implementation Notes.}
All metrics are computed on motion features extracted from expression, jaw, and pose parameters separately.  
Fréchet-based metrics use covariance statistics estimated from entire test sequences,  
and SID diversity follows the same clustering configuration as in the DualTalk benchmark for comparability.  
These metrics provide complementary views of realism, synchrony, and diversity in generated 3D conversational motion.

\input{figs/user_study_ui}

\section{User Study Details}\label{app:user-study}
We recruited 10 participants and randomly sampled 50 dialogue clips, including 25 from the \emph{test} dataset and 25 from the \emph{out-of-distribution (OOD)} dataset. 
Each participant evaluated all 50 clips, resulting in 500 pairs. 
For each clip, both agents were rated on four criteria, yielding 2000 criterion-level paired scores in total.
As shown in \Cref{fig:user_study_ui}, the interface presents four synchronized videos per sample: Agent~A, Agent~B, the ground-truth Agent (reference only), and the interacting User. 
Agent~A and Agent~B correspond to TIMAR and DualTalk, while participants are blinded to method identities. 
Participants are instructed to score only Agent~A and Agent~B, using the ground-truth Agent solely as a visual reference for conversational consistency. 
The UI supports replay and playback-speed adjustment.
Participants provide 1--5 Likert ratings on four criteria: Motion Naturalness, Facial Expression Naturalness, Interaction Naturalness, and Lip-sync Accuracy. 
Higher scores indicate better perceptual quality.

\input{tabs/user-study}

For aggregation, we compute a per-clip overall score for each method by averaging its four criterion scores. 
A preference is assigned when one method achieves a higher averaged score; ties are counted as 0.5 for each method. 
We report the preference rate for TIMAR over DualTalk and compute 95\% confidence intervals via bootstrap resampling over the 500 pairs with 20{,}000 iterations, as summarized in \Cref{tab:user-study-pref}. 
The analysis code and anonymized annotation data are provided in the supplementary material for reproducibility.

\input{tabs/diffhead}

\section{Diffusion Head Scalability Study}
\Cref{tab:diffhead-full} investigates the scalability of the diffusion head by varying its hidden dimension $d_{m}$ and the number of residual blocks $K$.  
Results on the \emph{test dataset} reveal a consistent trend of performance enhancement as model capacity increases,  
suggesting that the diffusion-based formulation can effectively leverage additional depth and width when larger computational budgets are available.  

%% file: figs/f_speech--f_head.tex
\begin{figure}[t]
\centering

\begin{minipage}{0.4\textwidth}
\centering
\includegraphics[width=0.6\linewidth]{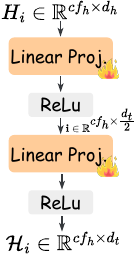}
\caption{
Architecture of the 3D head motion encoder
}
\label{fig:f_head}
\end{minipage}
\hfill
\begin{minipage}{0.3\textwidth}
\centering
\includegraphics[width=0.85\linewidth]{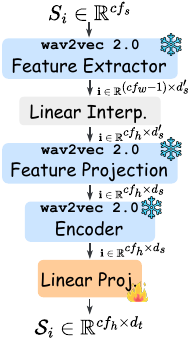}
\caption{
Architecture of the speech tokenizer
}
\label{fig:f_speech}
\end{minipage}

\end{figure}

%% file: figs/f_fusion--f_diff.tex
\begin{figure}[t]

\begin{minipage}{0.46\textwidth}
\centering
\includegraphics[width=\linewidth]{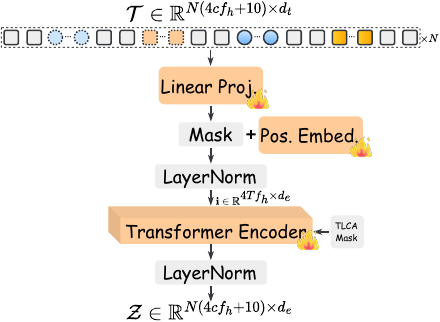}
\caption{
\textbf{Architecture of the Turn-Level Causal Multimodal Fusion module.}
Gray squares denote learnable separator tokens that delineate modality boundaries and turn transitions.
}
\label{fig:f_fusion}
\end{minipage}
\hfill
\begin{minipage}{0.48\textwidth}
\centering
\includegraphics[width=0.76\linewidth]{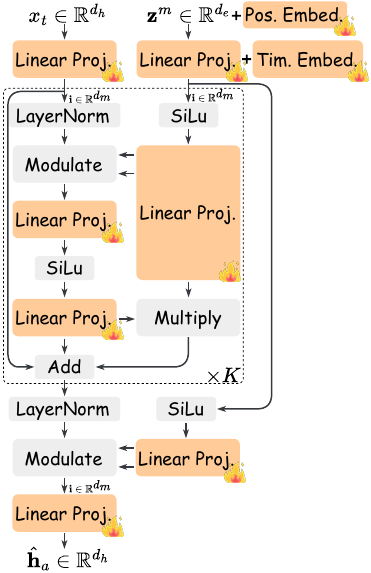}
\caption{
\textbf{Architecture of the Lightweight Diffusion Head.}
Multiple outgoing \textit{Linear Proj.} modules indicate that their outputs are chunked into several parts 
for modulation and gating operations within each residual block.
}
\label{fig:f_diff}
\end{minipage}

\end{figure}

%% file: tabs/dataset.tex
\begin{table}[t]
\centering
\setlength{\tabcolsep}{18pt}
\caption{
\textbf{Dataset statistics of the DualTalk benchmark.}  
The dataset comprises 50 hours of dual-speaker conversations with over 1000 unique identities.  
It includes official training, testing, and OOD splits, and the lower section reports the distribution of conversation rounds per sample.
}
\label{tab:dataset_stats}
\resizebox{0.5\textwidth}{!}{
\begin{tabular}{
>{\columncolor{gray!12}}l
c
}
\rowcolor{gray!40}\multicolumn{2}{l}{\textbf{\textit{Data scale}}} \\
Duration & 50h \\
Number of Identities & 1000+ \\
Number of All Samples &  5763 \\
Number of Training Samples & 4853 \\
Number of Test Samples & 533 \\
Number of OOD Samples & 377 \\
\specialrule{0em}{4pt}{4pt}
\rowcolor{gray!40}\multicolumn{2}{l}{\textbf{\textit{Distribution of conversation rounds}}} \\
1 Rounds & 1995~(34.6\%) \\
2 Rounds & 1126~(19.5\%) \\
3 Rounds & 1172~(20.3\%) \\
4 Rounds & 632~(11.0\%) \\
5 Rounds & 414~(7.2\%) \\
6+ Rounds & 424~(7.4\%)
\end{tabular}
}
\end{table}

%% file: figs/user_study_ui.tex
\begin{figure}
\centering
\includegraphics[width=\linewidth]{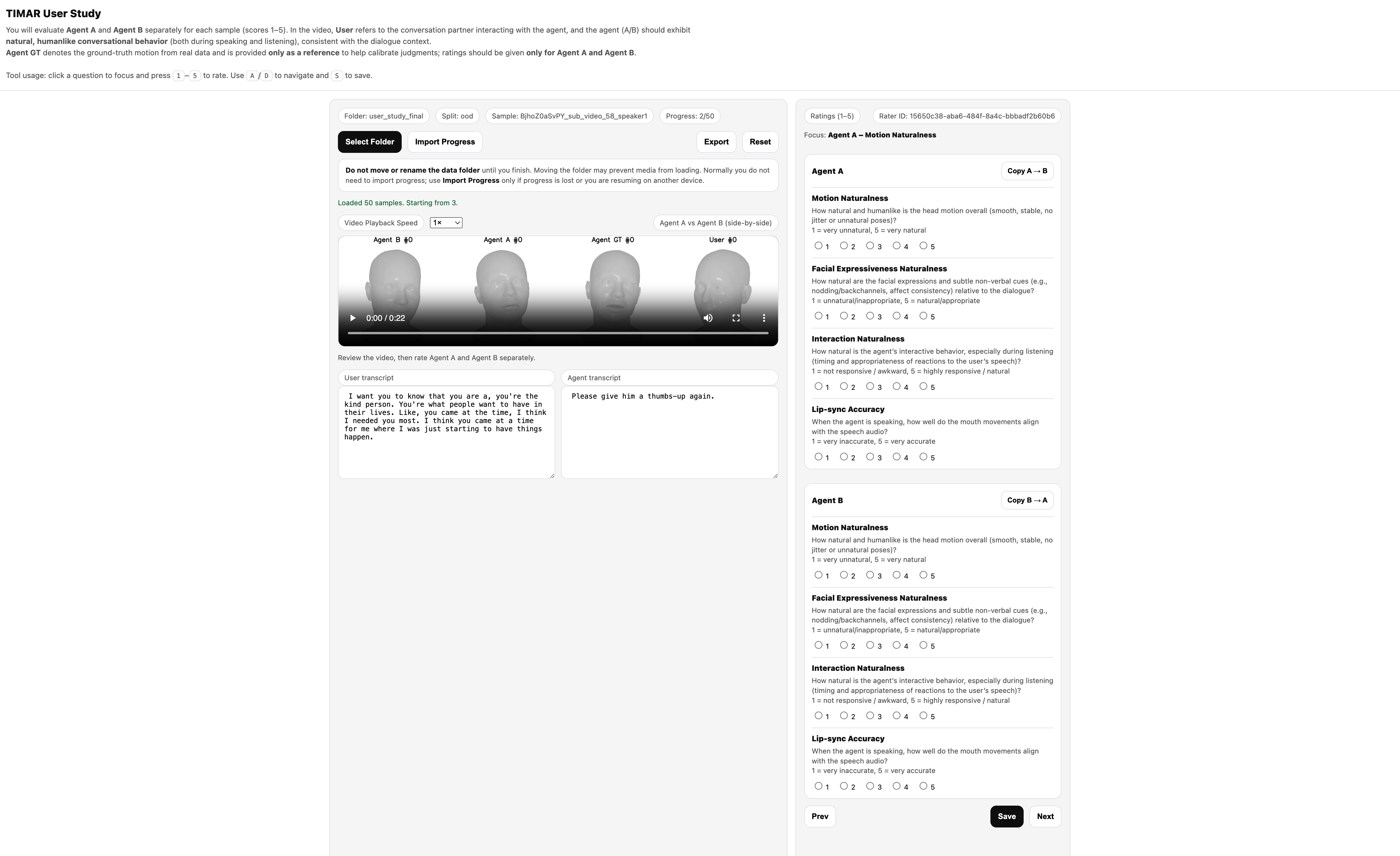}
\caption{
\textbf{User study interface.} 
For each sample, participants viewed Agent~A, Agent~B, the ground-truth Agent, and the interacting User side-by-side. 
Participants rated Agent~A and Agent~B independently on four criteria using 1--5 Likert scales.
}
\label{fig:user_study_ui}
\vspace{-2em}
\end{figure}

%% file: tabs/user-study.tex
\begin{table}[t]
\setlength{\tabcolsep}{6pt}
\caption{
\textbf{User study preference results.}
Preference rate denotes the proportion of votes for TIMAR over DualTalk with ties counted as 0.5.
Confidence intervals are 95\% bootstrap intervals computed over 500 pairs.
}
\label{tab:user-study-pref}
\centering
\resizebox{0.77\linewidth}{!}{
\begin{tabular}{lcccccc}
\rowcolor{gray!40}
\textbf{Criteria} & \textbf{$n$} & \textbf{TIMAR} & \textbf{DualTalk} & \textbf{Tie} & \textbf{Rate} & \textbf{95\% CI} \\
\specialrule{0em}{3pt}{3pt}

Motion Naturalness & 500 & 208 & 98 & 194 & 0.610 & [0.577, 0.642] \\
Facial Expr. Naturalness & 500 & 183 & 94 & 223 & 0.589 & [0.557, 0.620] \\
Interaction Naturalness & 500 & 161 & 88 & 251 & 0.573 & [0.543, 0.603] \\
Lip-sync Accuracy & 500 & 200 & 100 & 200 & 0.600 & [0.567, 0.633] \\
\rowcolor{gray!15}
Overall & 500 & 259 & 130 & 111 & 0.629 & [0.592, 0.666] \\

\end{tabular}
}
\end{table}

%% file: tabs/diffhead.tex
\begin{table*}[t]
\caption{
\textbf{Scalability study on Diffusion Head depth ($K$) and hidden dimension ($d_{m}$).}
We examine the influence of varying depth ($K$) and hidden dimension ($d_{m}$) on performance,  
with results reported on the \emph{test dataset}.  
Metrics with $\downarrow$ are better when lower (FD, P-FD, MSE, rPCC),  
and metrics with $\uparrow$ are better when higher (SID).    
}
\label{tab:diffhead-full}
\resizebox{\linewidth}{!}{
\begin{tabular}{
l
>{\columncolor{gray!12}}c
>{\columncolor{gray!12}}c
>{\columncolor{gray!12}}c
c
c
c
>{\columncolor{gray!12}}c
>{\columncolor{gray!12}}c
>{\columncolor{gray!12}}c
c
c
c
>{\columncolor{gray!12}}c
>{\columncolor{gray!12}}c
>{\columncolor{gray!12}}c
}
& \multicolumn{3}{c}{\textbf{FD} $\downarrow$} & \multicolumn{3}{c}{\textbf{P-FD} $\downarrow$}   & \multicolumn{3}{c}{\textbf{MSE} $\downarrow$} & \multicolumn{3}{c}{\textbf{SID} $\uparrow$}  & \multicolumn{3}{c}{\textbf{rPCC} $\downarrow$} \\
\rowcolor{gray!40}
\headcell{Methods}{} &
\headcell{EXP}{} &
\headcell{JAW}{$\times10^3$} &
\headcell{POSE}{$\times10^2$} &
\headcell{EXP}{} &
\headcell{JAW}{$\times10^3$} &
\headcell{POSE}{$\times10^2$} &
\headcell{EXP}{$\times10^1$} &
\headcell{JAW}{$\times10^3$} &
\headcell{POSE}{$\times10^2$} &
\headcell{EXP}{} &
\headcell{JAW}{} &
\headcell{POSE}{} &
\headcell{EXP}{$\times10^2$} &
\headcell{JAW}{$\times10^1$} &
\headcell{POSE}{$\times10^1$} \\
\specialrule{0em}{4pt}{4pt}
\rowcolor{gray!25}\multicolumn{16}{l}{$\mathbf{d_m=1024}$} \\
$K=1$ & 9.27  & 1.66  & 3.11  & 10.34 & 1.76  & 3.33  & 3.64  & 1.14  & 1.69  & 3.54  & 2.27  & 1.83  & 3.96  & 1.32  & 2.18 \\
$K=3$ & 8.97  & 1.57  & 3.08  & 9.93  & 1.65  & 3.28  & 3.58  & 1.07  & 1.61  & 3.53  & 2.35  & 1.85  & 4.12  & 1.22  & 2.18 \\
$K=6$ & 8.70   & 1.55  & 3.10   & 9.67  & 1.64  & 3.31  & 3.55  & 1.14  & 1.66  & 3.56  & 2.33  & 1.85  & 4.16  & 1.21  & 2.22 \\
$K=9$ & 8.63  & 1.66  & 3.02  & 9.62  & 1.75  & 3.23  & 3.55  & 1.16  & 1.64  & 3.57  & 2.31  & 1.85  & 4.29  & 1.34  & 2.19 \\
\specialrule{0em}{1pt}{1pt}
\rowcolor{gray!25}\multicolumn{16}{l}{$\mathbf{K=3}$} \\
$d_m=512$ & 9.31  & 1.73  & 3.09  & 10.34 & 1.82  & 3.3   & 3.72  & 1.19  & 1.69  & 3.54  & 2.31  & 1.85  & 3.91  & 1.48  & 2.11 \\
$d_m=768$ & 8.94  & 1.62  & 3.16  & 9.93  & 1.71  & 3.38  & 3.53  & 1.12  & 1.70   & 3.54  & 2.27  & 1.84  & 4.31  & 1.36  & 2.19 \\
$d_m=1024$ & 8.97  & 1.57  & 3.08  & 9.93  & 1.65  & 3.28  & 3.58  & 1.07  & 1.61  & 3.53  & 2.35  & 1.85  & 4.12  & 1.22  & 2.18 \\
$d_m=1280$ & 8.58  & 1.52  & 3.05  & 9.58  & 1.61  & 3.26  & 3.52  & 1.12  & 1.65  & 3.59  & 2.34  & 1.85  & 4.21  & 1.27  & 2.16 \\
\end{tabular}
}
\end{table*}